\documentclass{clv3}
\usepackage{hyperref}
\usepackage{xcolor}

\usepackage{amsmath}
\usepackage{booktabs}
\usepackage{CJKutf8}
\usepackage{subcaption}
\usepackage[toc,title,page]{appendix}

\DeclareMathOperator*{\argmax}{arg\,max}

\DeclareMathOperator*{\softmax}{soft\,max}

\definecolor{darkblue}{rgb}{0, 0, 0.5}
\hypersetup{colorlinks=true,citecolor=darkblue, linkcolor=darkblue, urlcolor=darkblue}

\runningtitle{A Tutorial on Word Alignment in the Age of Deep Learning}

\runningauthor{}

\begin{document}

\title{Word Alignment in the Era of Deep Learning: A Tutorial}

\author{Bryan Li\thanks{Philadelphia, PA. Email: \texttt{bryanli@seas.upenn.edu}}}
\affil{The University of Pennsylvania}

\maketitle

\begin{abstract}
The word alignment task, despite its prominence in the era of statistical machine translation (SMT), is niche and under-explored today. In this two-part tutorial, we argue for the continued relevance for word alignment. The first part provides a historical background to word alignment as a core component of the traditional SMT pipeline. We zero-in on GIZA++, an unsupervised, statistical word aligner with surprising longevity. Jumping forward to the era of neural machine translation (NMT), we show how insights from word alignment inspired the attention mechanism fundamental to present-day NMT. The second part shifts to a survey approach. We cover neural word aligners, showing the slow but steady progress towards surpassing GIZA++ performance.  Finally, we cover the present-day applications of word alignment, from cross-lingual annotation projection, to improving translation. 
\end{abstract}

\section{Introduction}

Word alignment is the task of identifying words in a source sentence that correspond to words in a target sentence, given that the sentences are translations of each other (see Figure~\ref{fig:wa_example}). It is a vital component of statistical machine translation (SMT) systems; however, in the age of modern neural machine translation (NMT), it has fallen out of prominence as an explicitly modeled task. Still, the concept of alignment motivated the development of the \textit{attention} mechanism~\cite{bahdanau2014neural}, which allows neural models to learn soft alignments between source and target sentences. Attention most prominently underlies the Transformer~\cite{vaswani2017attention}, the current state-of-the-art neural approach to natural language processing (NLP) problems, which extend the concept of alignment to within the same sequence -- self-attention. Still, the explicit word alignment task has become rather niche in the present-day literature.

The goal of our work is to argue for the importance, historical and continued, of the word alignment task. In the SMT age, word alignment was an essential preprocessing step which allowed decomposition of translating sentences into translating phrases. In the modern NMT age, word alignment remains as a useful, yet under-explored task. One of its main applications is for \textit{annotation projection}, a technique to extend datasets and tasks cross-lingually. Other applications are to improve translation, and as \textit{monolingual} word alignments to reframe NLP tasks such as text simplification. Finally, regardless of the era, (word) alignments are innately interpretable and thus useful in downstream analysis of a system's predictions.

\subsection{Target Audience}
This tutorial aims to be broadly useful to NLP researchers at different levels. The primary audience are those researchers who have grown up in the deep learning era -- such as early-stage PhD students -- and may not as keen on the word alignment task.
For more senior researchers, we provide an overview of the word alignment field from the 1990s to the early 2020s. We further situate the task in the context of the attention-based models so prevalent in modern-day NLP. In both covering the basic concepts and providing intuition on the details, we hope that our tutorial is of general interest to anyone interested in language technology.

\subsection{Key concepts}

\begin{figure}[t]
    \centering
    \includegraphics[width=.65\textwidth]{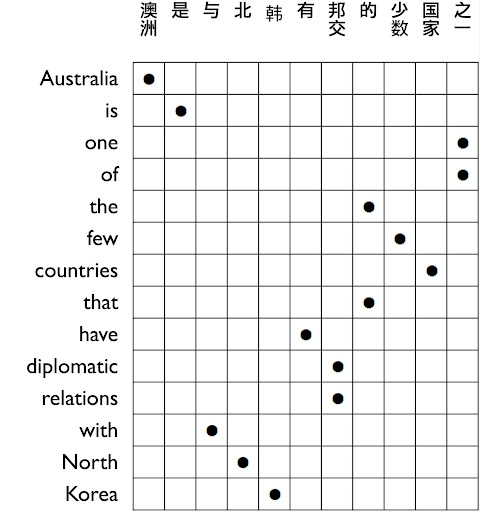}
    \caption{Example word alignment between English and Chinese. Dotted cells represent where an alignment occurs. For example, the top left cell shows that ``Australia'' and ``\begin{CJK*}{UTF8}{gbsn}澳洲\end{CJK*}'' are aligned.}
    \label{fig:wa_example}
\end{figure}

As stated previously, \textbf{word alignment} is the task of identifying words in a source sentence that correspond to words in a target sentence, given the sentences are translations of each other. \textbf{Machine translation} (MT) is the task of developing systems that can translate source language text into target language text.

On the data side, \textbf{sentence pair} consists of a source sentence and a target sentence. A dataset of sentence pairs is called a \textbf{bitext}, or a \textbf{parallel corpus}.

Word alignments can be a) one-to-one, b) one-to-many, c) many-to-one, or d) many-to-many. Figure~\ref{fig:wa_example} depicts an example of a word alignment between a Chinese and English sentence pair. It depicts one-to-one (``Australia'' to ``\begin{CJK*}{UTF8}{gbsn}澳洲\end{CJK*}'') and many-to-one
(``diplomatic'', ``relations''  to ``\begin{CJK*}{UTF8}{gbsn}邦交\end{CJK*}'') alignments. We hope the observant reader can imagine examples of the other cases.

The final key concept is the title phrase ``the era of deep learning.'' Let us decompose this phrase into its parts. 
\textbf{Deep learning} refers to machine learning methods based on multi-layered (hence ``deep'') neural networks. \textbf{Neural networks} are computer systems inspired by the operation of neurons in the human brain.
While neural networks have existed since the mid-20th century, it was only until the 2000s that the convergence of developments in both computational power and neural modeling approaches allowed their successful application to vision and language problems. 
Thus, \textbf{the era} of deep learning refers to 2000s to the present. In this era deep learning is the go-to approach towards tackling problems in artificial intelligence, and statistical methods are often seen as antiquated. Crucially though, unsupervised word alignment retains a strong statistical baseline, GIZA++ that remains competitive with neural aligners up to the late 2010s. It is in this context that we situate our current tutorial.

\subsection{Structure of Tutorial}
The sections of this tutorial follow in a roughly chronological order.
We will provide mathematical formulations and intuitive definitions of concepts throughout. We intend that each section is fairly self-contained, and readers can read the sections in any order. 

Broadly speaking, this tutorial is structured in two parts. Sections~\ref{sec:smt} to~\ref{sec:att} take a more in-depth tutorial approach, whereas Sections~\ref{sec:neural_wa} to~\ref{sec:app_wa} take a more comprehensive literature survey approach.

Section~\ref{sec:smt} introduces \textbf{statistical machine translation} (SMT). We first provide mathematical formulations of the SMT task, and the relate it to the word alignment task. We then describe the operation of Moses, typical phrase-based SMT system, at a high-level. This will elucidate the important role word alignment plays in this pipeline.

Section~\ref{sec:stat_wa} describe \textbf{statistical approaches to word alignment}. These are generally unsupervised approaches. We summarize the famous IBM family of aligners and related work. Then we zero-in on GIZA++, a notable parameterization of the IBM models.

Section~\ref{sec:nmt} introduces \textbf{neural machine translation}. We focus on encoder-decoder models, one popular approach to NMT. We show that this end-to-end formulation of MT does not concern itself with word alignments.

Section~\ref{sec:att} describes \textbf{the attention mechanism}, which allows neural models to focus on parts of the input sequence when generating its output. Attention extends encoder-decoder models to perform a soft-alignment between target and source tokens. To understand how well attention correlates to, and differs, from word alignments, we review a few works in this direction.

Section~\ref{sec:neural_wa} performs a comprehensive survey on \textbf{neural approaches to word alignment}. We categorize these models into three broad approaches: induction from attention, unsupervised, and guided. We also discuss work that extracts alignments from multilingual neural language models, and also classify them by these three approaches.

Section~\ref{sec:app_wa} performs a broadly scoped survey on the \textbf{applications of word alignment}. We focus on approaches more relevant to modern-day NLP, covering a few key works from the statistical era. We find that a major use-case of word alignment is in annotation projection, extending tasks and datasets cross-lingually. We also discuss how word alignment can improve NMT and computer-assisted translation.
\section{Statistical Machine Translation and Word Alignment}
\label{sec:smt}

\subsection{Formalizing Statistical Machine Translation}
\label{ssec:smt}
Machine translation is the task of translating a source sentence $F$ into a target sentence $E$, where
\[ F=f_1,...,f_m \qquad\qquad E=e_1,...,e_n \]

As a mnemonic, suppose you know English but not French---you would want to translate source \textit{F}rench into target \textit{E}nglish. \textbf{Statistical machine translation} systems create a model for the probability of every target sentence $E$ given some source sentence $F$. They find $\hat{E}$, the hypothesis which maximizes the probability

\begin{equation}
\label{eq:smt}
    \hat{E} = \argmax_E Pr(E | F; \theta) 
\end{equation}

where $\theta$ are the (learned) parameters of the model and capture the probability distribution. By applying Bayes's theorem, we can decompose this equation:
\begin{align*}
    \hat{E} &= \argmax_E Pr(E | F; \theta) \\
    &= \frac{\argmax_E Pr(F | E; \theta) Pr(E)}{Pr(F)}
\end{align*}
\begin{align}
    \hat{E} &\propto \argmax_E Pr(F | E; \theta) Pr(E)
\end{align}

$Pr(F | E; \theta)$, or simply $Pr(F | E)$, is the \textbf{translation model}---given this target sentence, how likely is the source sentence . $Pr(E)$ is the \textbf{language model} (LM)---given this target sentence, how fluent is it in the target language. SMT systems require both \textbf{bitext} data to learn the translation model, as well as \textbf{monolingual} data in the target language to learn the language model.

\citet{Neubig2017NeuralMT} specifies the three problems a good translation system must address:
\begin{enumerate}
    \item \textit{Modeling}: How will we model $P(E | F; \theta)$, what are its parameters $\theta$, and how do the parameters specify a probability distribution?
    \item \textit{Learning}: How will we learn the parameters $\theta$ from the training data?
    \item \textit{Search}:  How will we find, or \textbf{decode}, the most probable sentence?
\end{enumerate}

\subsection{Formalizing Word Alignment}
The word alignment task was directly motivated by the SMT task. An SMT system aims to model $Pr(F | E)$, but to do so over all tokens $Pr(f_1, ...,f_m | e_1, ..., e_n)$ is difficult. Word alignments were thus introduced as a set of hidden variables $a=a_1,...a_m$ to make the problem more tractable. Suppose we have a given alignment $a$. Then we have the problem
\begin{equation*}
    Pr{(f_1,\ldots,f_m,a_1,\ldots,a_m| e_1,\ldots,e_m)} = Pr{(F,a| E)}
\end{equation*}

By summing over the possible alignments, we recover the original translation model:
\begin{equation}
\label{eq:wa_smt}
    \sum_{a}{Pr{(F,a | E)}} = Pr{(F| E)}
\end{equation}
Essentially, word alignment decomposes the task of translating an entire sentence into translating parts of it.

\citet{och2003systematic} present the following formalization of word alignment. Given a source string  $F=f_1, ..., f_j, ..., f_m$ and a target string $E=e_1,..., e_i, ..., e_n$, an alignment $\mathcal{A}$ is a subset of the Cartesian product of word positions:
\begin{equation}
    \mathcal{A} \subseteq \{(j,i): j=1, ..., m ; i = 1, ..., n\}
\end{equation}

\begin{figure}[t]
    \centering
    \includegraphics[width=.65\textwidth]{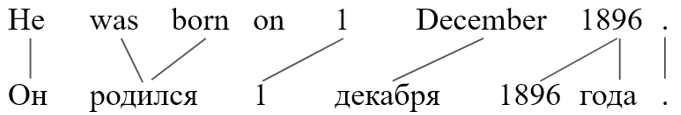}
    \caption{Example Russian to English word alignment. The mapping and relations are respectively \\ $\mathcal{A}=\{(1,1), (2,2),(2,3), (3,4), (4,5), (5,6),(6,6),(7,7)\}$ \\ $a: a_1=1,a_2=2,a_2=3, a_3=4,a_4=5,a_5=6,a_6=6,a_7=7$}
    \label{fig:example_2}
\end{figure}

$\mathcal{A}$ is a mapping of individual alignment points $a_j$, each of which maps a source index \textit{j} to a target index \textit{i}. We can view $a$ as a relation where $a_j=i$. For example, Figure~\ref{fig:example_2} gives both the mapping and relation views of word alignment.

This complete, albeit general formulation allows for an exponential search space of alignments. In practice, alignment models often impose additional constraints to make the problem more tractable and produce better quality alignments. One common restriction is to enforce an injection: each source word maps to exactly one target word. A null word $e_0$ is added to the set of target indices to maintain this property when some source word has no alignment. Note that this restriction preventing one-to-many and many-to-many mapping, but this issue is often addressed by running word alignments in both directions and combining the two.

Recall Equation~\ref{eq:wa_smt}, which shows the relationship between the translation model and word alignment. Given a a corpus of sentence pairs $(F_1,E_1),...,(F_s, E_s)$, this model is tasked to approximate the model parameters $\theta$ by maximizing the probability
\begin{equation}
    \hat{\theta} = \argmax_{\theta} \prod_{i=1}^{s} \sum_{a} Pr(F_i,a|E_i;\theta)
\end{equation}

\subsection{Tokenization and Vocabulary}
\label{ssec:tok_vocab}
Let us cover some additional key concepts before moving on. \textbf{Tokenization} is the process of splitting a text into individual tokens. Most simply, a token could be a word; other approaches operate at the subword-level, or even the character-level. This is a preprocessing step for all machine translation systems, statistical or neural.

A \textbf{vocabulary} is the set of words (or in our case, tokens) in a language. 
For machine translation, we will have both a source vocabulary and a target vocabulary. We define the \textbf{vocabulary size} to be the number of tokens in a given vocabulary. This is generally much smaller than the number of valid words in a natural language. For words that fall outside the vocabulary, NLP systems generally map them to an unknown token \textit{<unk>}.

\subsection{Phrase-Based Machine Translation}
\label{ssec:phrase_smt}

We use the Moses~\cite{koehn-etal-2007-moses} open-source toolkit as an exemplar of a typical SMT system. Moses is a phrase-based machine translation system; as the name implies, this works by translating source phrases, i.e., parts of a sentence, then recombining these target phrases into fluent target sentences. 
As a translation system, Moses consists of 3 parts: the language model, the translation model, and the decoder. In this section, we describe aspects of Moses relevant to word alignment---namely the translation model. Further details are given in Appendix~\ref{appsec:moses}.

The generation process of phrase-based MT occurs in three stages:
\begin{enumerate}
    \item \textit{Partition}: Given a source sentence, create all of its possible partitions, where each partition is a set of source phrases.
    \item \textit{Lexicalization}: Given a partition, translate its source phrases into hypothesized target phrases.
    \item \textit{Reordering}: Given a set of target phrases, permute it into a grammatical sentence in the target language. This involves rearranging words and moving them across phrases boundaries.
\end{enumerate}

\begin{figure}[ht]
    \centering
    \includegraphics[width=.4\textwidth]{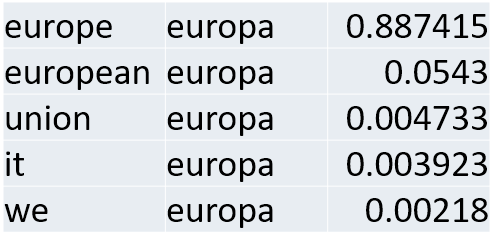}
    \caption{An excerpted section of a phrase translation table, where source is English (En), and target is German (De). The probabilities are $Pr(\text{En}|\text{De})$. In this excerpt, both source and target phrases are of length 1, but in general, entries can be longer than 1 token.}
    \label{fig:t_table}
\end{figure}

Word alignment is evidently relevant in the partition and lexicalization stages. Let us now describe word alignment's role in SMT. As a preprocessing step, we first learn word alignments from a large parallel corpus\footnote{For example, the United Nations Parallel Corpus (\url{https://conferences.unite.un.org/uncorpus}) is a set of parallel corpora in six languages. The fr-en corpus consists of 25 million sentence pairs.}. As for how they are learned, we describe one such word aligner, GIZA++, in Section~\ref{ssec:giza}. We apply the trained aligner to the entire training corpus, and for each source phrase, accumulate all the target phrases\footnote{To be specific, GIZA++ enforces an injection, so will only have one-to-one and many-to-one alignments, where one is a word and many is a phrase. But by training aligners in both directions and combining the two, we can have all 4 types of alignments, and thus align source and target phrases.} it maps to. Converting counts to probabilities, we have phrase translation tables (t-tables), such as is shown in Figure~\ref{fig:t_table}.

We can use the source phrase entries to \textit{partition} a source sentence, and \textit{lexicalize} given this partition into the associated target phrase(s). For any given sentence, there is a large state space of possible partitions multiplied by each possible target translation. Therefore, it is the job of the Moses MT system to efficiently choose the best, or n-best, translations. These details go beyond the present tutorial, of course.

As further evidence for the key role of word alignment to statistical machine translation, \citet{callison-burch-etal-2004-statistical} find that modifying the EM algorithm for GIZA++ to incorporate a small amount of manual word alignments, in addition to those automatically discovered word alignments, improves both translation and alignment performance of the downstream SMT system.
\section{Statistical Approaches toward Word Alignment}
\label{sec:stat_wa}

Section~\ref{ssec:phrase_smt} shows that word alignment occupies a key stage in SMT, in that it decomposes the harder task of translating entire source sentences to the simpler task of translating source phrases. We now describe one statistical word alignment system.

\subsection{GIZA++: A Strong Baseline for Word Alignment}
\label{ssec:giza} 
GIZA++~\cite{och2003systematic} is an unsupervised word alignment tool with surprising longevity, remaining the state-of-the-art well into the 2010s, and still being cited today. GIZA++ serves as an implementation of the IBM family of word alignment algorithms~\cite{brown1993mathematics}. These are numbered in order of increasing complexity. Originally, there are (IBM) Models 1-5, and \citet{och2003systematic} introduces Model 6 as the default GIZA++ model. Other popular statistical word alignment tools are fast\_align~\cite{dyer-etal-2013-simple}, and Berkeley Aligner~\cite{liang-etal-2006-alignment}. These are discussed briefly in Appendix~\ref{appsec:swa}.

GIZA++ is an unsupervised word alignment tool, which works by leveraging language-independent statistical methods. The approach of GIZA++, and the IBM models in general, is to view word alignment as a hidden variable in the translation process. Statistical estimation is used to compute the ``optimal'' model parameters, and alignment search is performed to compute the best word alignment. Each IBM model models $Pr(F,a|E;\theta)$ differently.

In the remainder of this section, we'll first walk through IBM Model 1 in detail. We'll then cover, at a higher-level, the additional modeling assumptions made by the GIZA++ model.

\paragraph{``Unsupervised'' Word Alignment} As an aside, unsupervised in the context of word alignment means learning this task without any labeled word alignment data. We do need some supervision from human annotators in collecting parallel sentences. Still, this is much less of an involved task that finding and training annotators to perform word alignment. Parallel sentences can be found as artifacts from international organization proceedings, or can be written by readily available human translators.

\subsection{IBM Model 1}\leavevmode\newline
IBM Model 1 is the simplest model and makes the fewest assumptions. It has a uniform prior over the possible alignments, which means every alignment is equally plausible. The translation probability for a given alignment $a$ is specified by
\begin{equation}\label{eq:model1}
    Pr(F,a|E) = \frac{Pr(m|n)}{(n+1)^{m}} \cdot \prod_{j=1}^{m} Pr(F_j|E_{a_j}) ,
\end{equation}
where $m, n$ are the lengths of $F$ and $E$ respectively.

Let us explain the constituents of this formula. The numerator, $Pr(m|n)$ is the probability of choosing a source length given a target length. The denominator, $(n+1)^m$ specifies the uniform prior over all possible alignments; it is raised to the power $m$, as we iterate over $m$ source tokens. For each source index \textit{j} we have $Pr(F_j|E_{a_j})$, the probability of source token $F_j$ given its predicted aligned target word $E_{a_j}$. We take the product of these probabilities to translate all source phrases, and then multiply this product by a normalizing factor.  

A system implementing IBM Model 1 learns over a large parallel corpus. For each sentence pair, it needs to learn both a) the best alignment, and b) the translation probabilities. This is an unsupervised approach, so both must be learned in parallel, leading to a very natural application of the EM algorithm. We refer interested readers to~\citet{och2003systematic}.

\subsection{IBM Model 6}

The GIZA++ model, IBM Model 6, is a combination of IBM Model 4, and the HMM aligner of \citet{vogel-etal-1996-hmm}.

\paragraph{HMM Aligner} This decomposes the alignment probability into three different probabilities: length probability, prior alignment probability, and lexicon probability. Let us focus on the prior alignment probability. The HMM aligner uses \textit{locality in the source language} --- when aligning a source word $f_i$, it considers prior alignments $f_0,...,f_{i-1}$. In fact, we only need to consider $f_{i-1}$ given the Markov assumption that the future is independent of the past given the present. 

\paragraph{IBM Model 4} This extends Model 1 with more assumptions, namely using \textit{fertility} and \textit{locality in the target language}. 
The fertility $\phi$ of a target word $e_i$ is simply the number of aligned source words. The model learns the probability that $e_i$ is aligned to $\phi$ words.
An example of the usefulness of fertility is the German word ``\"ubermorgen,'' which translates to the four English words ``the day after tomorrow''\footnote{This example from~\citet{och2003systematic} has as the source language English and the target German.}.

For locality in the target language, every word depends on the previous aligned word. A further refinement of Model 4 is in considering the word classes\footnote{Derivation of word classes, in the Moses pipeline, are described in Appendix~\ref{appsec:moses}.} of the surrounding words.

In sum, Model 6 is a log-linear combination of Model 4 and an HMM aligner, to make use of locality in both the source and target languages. As with Model 1, Model 6 is also learned through the EM algorithm, but takes quite a bit longer to train due to its additional complexity.

To close this section, we quote~\citet{zhao-gildea-2010-fast}, ``IBM Model 4 is so complex that most researchers use GIZA++...and IBM Model 4 itself is treated as a black box.'' Still, we hope the quick overview we have presented gives readers a sense as to the statistical effort placed behind GIZA++.

\section{The Age of Neural Machine Translation}
\label{sec:nmt}
Sections~\ref{sec:smt} and~\ref{sec:stat_wa} have shown how the word alignment task is intricately linked statistical machine translation. These sections have also shown how such statistical approaches require significant \textit{feature engineering}, designed with a high-level of statistical rigor, and further tuned by well-trained researchers. In contrast, \textbf{neural machine translation} (NMT) utilizes neural networks to learn sophisticated functions for language modeling, abstracting away the hard work of feature engineering (and introducing another line of hard work!). For a review on how neural networks work, see Appendix~\ref{appsec:nn}.

In the age of deep learning, neural machine translation has almost entirely supplanted statistical machine translation. Word alignment as an explicit task is no longer necessary, and thus research using word alignments has become rather niche. 

In this section, we first describe the encoder-decoder model paradigm commonly used in present-day NMT. Next we cover the attention mechanism, which incorporates into this paradigm a form of alignment. We then review some work analyzing the correlation between attention weights and word alignment, concluding while they share some similarities, they are not interchangeable.

\subsection{Encoder-Decoder Models}
\label{ssec:enc_dec}
\begin{figure}[t]
    \centering
    \includegraphics[width=.8\textwidth]{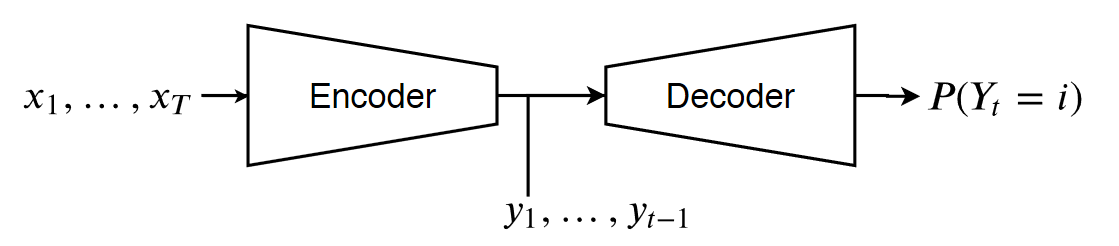}
    \caption{A encoder-decoder model for a seq2seq task, excerpted from~\citet{ippolito2022nlg}. Note the two inputs to the decoder at time step $t$: a) the encoder output, and b) the past decoder outputs.}
    \label{fig:enc_dec}
\end{figure}

Here we explain a basic encoder-decoder model~\cite{sutskever2014sequence}. Let us generalize from the machine translation problem (from $F$ to $E$) to a sequence-to-sequence problem (from $x$ to $y$). A \textbf{encoder-decoder} model consists of two neural networks. 
An example encoder-decoder model is shown in Figure~\ref{fig:enc_dec}. 

The encoder takes in the input, and the decoder decodes the output. Most of the time, the encoder and decoder share very similar model architectures. The difference is in their input and output dimensions.

An \textit{encoder} takes as input $x_1,...,x_j$\footnote{Neural models actually operate on embeddings, which are real-valued vector representations for words. For MT, we have both source embeddings and target embeddings. We omit this detail for simplicity of notation, but keep it in mind.} and outputs a hidden-state representation. In the basic model, this is a fixed-length, real-valued vector $h^{(x)}$.

The \textit{decoder} proceeds in time steps, decoding 1 word at a time. At each $t$, it takes two inputs a) $h^{(x)}$ and b) the prior predicted words $\hat{y}_1,...,\hat{y}_{t-1}$, and outputs an probability distribution over the target vocabulary $Pr(y_t=i)$\footnote{The decoder actually outputs another embedding. $Pr(y_t=i)$ is the softmax of the output embedding times the target embedding matrix --- i.e., how similar is each target embedding to the output.}.


In mathematical notation, the decoder defines the probability of an output sequence $y$
\begin{equation}\label{eq:dec}
    Pr(y) = \prod^j_{t=1} Pr(y_t | \{y_1,...,y_{t-1}\}, h^{(x)})
\end{equation}

\begin{figure}[t]
    \centering
    \includegraphics[width=.65\textwidth]{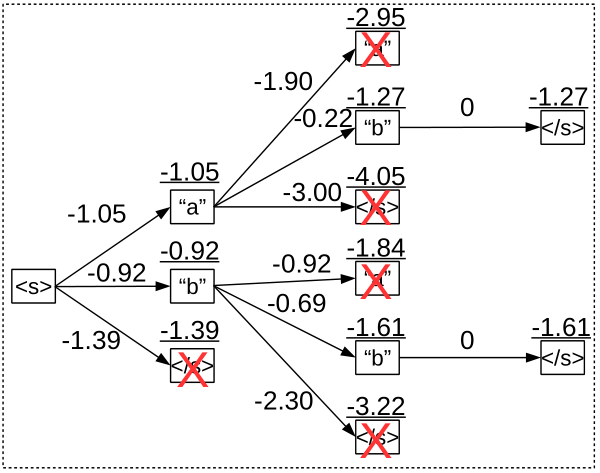}
    \caption{A visualization of beam search with $b=2$, excerpted from~\citet{Neubig2017NeuralMT}. The numbers labeling arrows are log probabilities transitioning between nodes, while numbers labeling nodes are the sum of log probabilities to this node. Red `X`s indicates the pruned hypotheses at each time step.}
    \label{fig:beam_search}
\end{figure}

Given these $t$ probabilities, we would like to convert them to target tokens. We do so using a \textit{sampling algorithm}. Most obviously, we can greedily select the highest probability vocabulary item at each time step $t$, and concatenate these together. A more informed approach is to use \textbf{beam search}. At each time step, we consider the $b$ best hypotheses, where $b$ is the size of the beam. At the $i$th time step our hypotheses are of length $i$. At each successive step, we select the next $b$ best hypotheses (word length $i+1$) and prune the rest. At the last step, we pick the best of the $b$ remaining hypotheses. Figure~\ref{fig:beam_search} shows a visualization of the beam search process.

An encoder-decoder model is a general paradigm that can be applied to various text to text tasks --- as we do to translation in this tutorial. The underlying neural network architecture can be recursive neural networks (RNNs), transformers, or anything else.

\section{Attention: Implicit (Word) Alignments?}
\label{sec:att}
Whereas Sections~\ref{sec:smt} and~\ref{sec:stat_wa} show how statistical MT and word alignment are intrinsically linked, Section~\ref{sec:nmt} shows how neural MT methods are can perform end-to-end translation of full sentences, without any explicit word alignments.

In this section, we guide readers through the \textbf{attention} mechanism, a method first developed to incorporate a notion of alignment back into NMT systems. After formalizing attention with mathematical notation, we review several works analyzing the correlation of attention weights with word alignments. We find attention implicitly but imperfectly models alignments, underscoring our argument towards the continued importance of the word alignment task today.

\subsection{Attention}
\textbf{Attention} was introduced in ``Neural Machine Translation by Jointly Learning to Align and Translate''~\cite{bahdanau2014neural}. The authors propose attention as a method for aligning words between the input and output sentences. They term these ``soft-alignments'' because each target word has a set of real-valued weights, one for each source word. In contrast, word alignments have source and target words being either aligned or not (0 or 1). Still, the intuition remains that the general concept of aligning words between source and target sentences can also inform the decision of neural MT systems. In fact, \citet{bahdanau2014neural} mention ``attention'' only 3 times, but mention ``alignment'' over 20 times.

Let us momentarily pause to describe the motivation for attention. The basic encoder-decoder model has a glaring issue, in that the encoder representing sequences of all lengths into a fixed-length vector. For especially long sequences, the encoder would generate a very dense representation, and the decoder will likely have difficulty decoding all pieces of information.

Attention solves this issue within the encoder-decoder paradigm. An attentional model will keep all encoded vector representations of input tokens, and reference these in the decoding step. At each decode time step $t$, then, we have access to each part of the input (as well as the prior decoded tokens). Intuitively, the decoder has to pay attention to different parts of the input when making each decision.

\paragraph{Attention Explained}
We now provide the mathematical formulation of the original attentional model~\citet{bahdanau2014neural}, which extends an RNN-based encoder-decoder model. As in prior notation, let $f$ represent the source sentence and $e$ represent the target sentence. Let $f_j$ be the source word at index $j$, and and $e_i$ be the target word at index $i$.

\paragraph{Encoder} To encode the source sentence, we use two RNNs, one in the forward (left-to-right) direction, and one in the backward (right-to-level) direction:
\begin{gather}
    \overrightarrow{h}_j^{(f)} = RNN(embed(f_j), \overrightarrow{h}_{j-1}^{(f)}) \\
    \overleftarrow{h}_j^{(f)} = RNN(embed(f_j), \overleftarrow{h}_{j+1}^{(f)})
\end{gather}
where $\overrightarrow{h}_j^{(f)}$ and $\overleftarrow{h}_j^{(f)}$ are the RNN hidden states.

Each source word is then represented bidirectionally as the concatenation of forward and backward vectors
\begin{equation}
    h_j^{(f)} = [\overleftarrow{h}_j^{(f)} ; \overrightarrow{h}_j^{(f)}] .
\end{equation}
The source is now encoded as the set of vectors $h_0^{(f)}, ..., h_n^{(f)}$
\paragraph{Decoder}
Recall Equation~\ref{eq:dec}:
\begin{equation*}
    Pr(y) = \prod^j_{t=1} Pr(y_t | \{y_1,...,y_{t-1}\}, h^{(x)})
\end{equation*}

In the original model, the same vector $h^{(x)}$ is used at all time steps. In contrast, the attentional model uses a different context vector $c_t$ at each time step $t$. Maintaining the prior notation, and noting that $y$ is equivalent to $f$,

\begin{equation}
    Pr(y) = \prod^j_{t=1} Pr(y_t | \{y_1,...,y_{t-1}\}, c_t) .
\end{equation}

$c_t$ is defined as the sum of encoder hidden states weighted by alignment scores $\alpha_t$
\begin{equation}\label{eq:context}
    c_t = \sum_{j=1}^m \alpha_{t,j} ,
\end{equation}
where $c_t$ is of dimension $m$, and thus defines a weight distribution over the input words. $\alpha_t$ is the attention vector.

Now we have seen where $\alpha_t$ is used. To explain how $\alpha_t$ is calculated, let us work forwards from the decoder hidden state, specified by
\begin{equation}
    h_t^{(e)} =  Enc([embed(e_{t-1});c_{t-1}], h_{t-1}^{(e)}).
\end{equation}
We see that the current decoder state takes in the prior context vector $c_{t-1}$ and the prior decoder hidden state $h_{t-1}^{(e)}$

We then calculate an attention score $a_t$, where each of its element $a_{t,j}$ is
\begin{equation}
    a_{t,j} = \text{Attn}(h_j^{(f)}, h_t^{(e)}) .
\end{equation}
$\text{Attn}(\cdot)$ is an arbitrary function that takes in two vectors and outputs a weight for how much we should focus on a particular input encoding at the current decode time step. \citet{bahdanau2014neural} use a simple RNN:
\begin{equation}
    \text{Attn}(h_j^{(f)}, h_t^{(e)}) = v_a^\top \tanh{(W_a[h_j^{(f)}; h_t^{(e)}])}
\end{equation}
Where $W_a$ is the weight matrix of the first layer and $v_a^\top$ is the vector of the second layer.

Finally, to use the attention vector as a probability distribution, we apply softmax
\begin{equation}
    \alpha_t = \softmax(a_t),
\end{equation}
so that the weights sum to 1. This is then used in Equation~\ref{eq:context} from above.

\begin{figure}
    \centering
    \includegraphics[width=.33\linewidth]{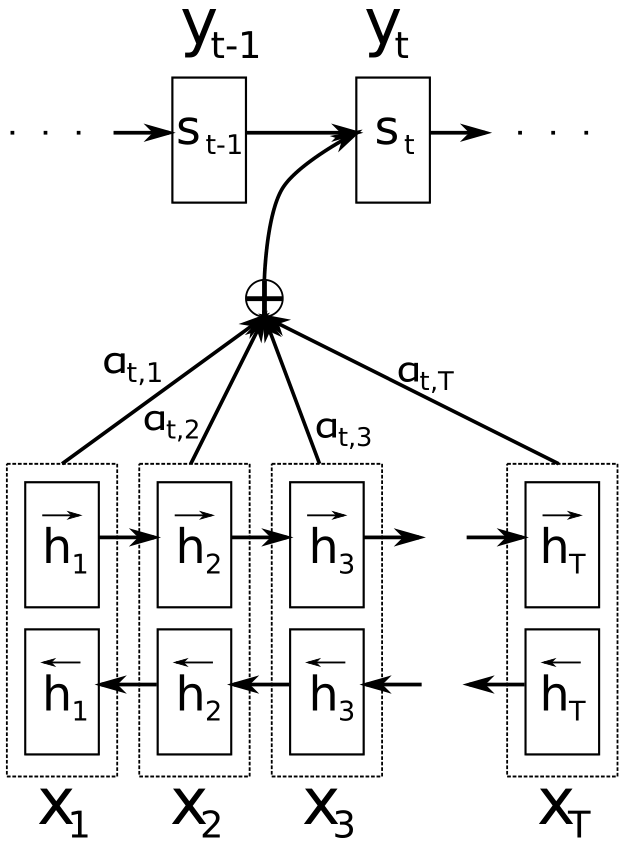}
    \caption{An illustration of the attentional NMT model, when generating the target word $y_t$ given the source sentence $x_1, x_2, x_3, ..., x_T$. Reproduced from~\cite{bahdanau2014neural}.}
    \label{fig:attention_model}
\end{figure}

For each time step $t$, we now have a representation consisting of a context vector $c_t$ and a decoder hidden state $h_j^{(e)}$. We can use these, for instance, to calculate a softmax distribution over the next word(s):
\begin{equation}
    Pr(e_t) = \softmax(W_{hs} [v_j^{(e)};c_t] + b_s)
\end{equation}
We can then use the sampling algorithms described in Section~\ref{ssec:enc_dec} to decode the target language tokens. The attentional NMT model is shown in Figure~\ref{fig:attention_model}.

\subsubsection{Attention Functions and Transformers}
\label{sssec:att_trans}
It turns out that attention is a highly effective and influential approach, beyond machine translation and to the NLP field as a whole. We now summarize follow-up work in the field.

\citet{luong-etal-2015-effective} proposes several modifications to the original attention mechanism. First, they use a unidirectional encoder, instead of bidirectional. Second, to calculate attention, they consider the decoder state at $t$, instead of $t-1$, allowing us to directly use $h_t^{(e)}$ without the extra RNN layer. Finally, they introduce several different attention functions, the simplest (and most effective) of which is dot-product attention:
\begin{equation}
    \text{Attn}(h_j^{(f)}, h_t^{(e)}) = {h_j^{(f)}}^\top \cdot h_t^{(e)}
\end{equation}
The main takeaway here is that regardless of differences in formulation, attention is quite powerful as a general mechanism.

Attention solves one issue with RNN-based encoder-decoder models, but another major issue remains. RNNs are slow, during both training and inference, because generating a hidden state at time step $t$ requires us to have generated all prior time steps $0,...,t-1$. \citet{vaswani2017attention} propose the \textit{transformer}, which removes recurrence entirely from the language model, in favor of only using attention\footnote{For $\text{Attn}(\cdot)$, \citet{vaswani2017attention} use scaled dot-product attention, which adds a scaling factor $\sqrt{n}$ to dot-product attention.}.

The key idea of the transformer is the idea of \textit{self-attention}, which turns attention in on itself. The attention described above relates words between two different sequences. Self-attention operates in much the same way, except we now consider a single sequence $s$,  learning correlations between each word $s_i$ with the prior words $s_0,...,s_{i-1}$. A transformer thus uses three instances of attention: source self-attention, target self-attention, and source-target attention. Further contributions of the transformer can be found in Appendix~\ref{appsec:trans}.

Today, the transformer architecture is the dominant approach to NLP tasks. Notable follow-up work includes BERT~\cite{devlin-etal-2019-bert} and GPT~\cite{gpt}, among many others.

\subsection{Attention and Alignment}
\label{sssec:att_align}
A major advantage of attention, aside from its improvements to language models' peformance, is in its interpretability. We now briefly summarize the body of work investigating the correspondence of attention with ``traditional'' word alignment.

\begin{figure}[t]
    \centering
    \includegraphics[width=.6\linewidth]{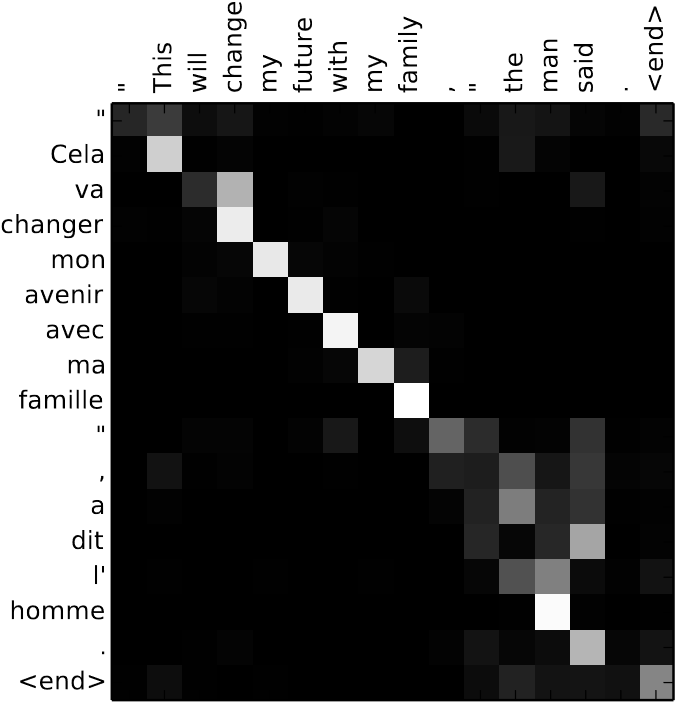}
    \caption{A matrix visualization of alignments produced by an attentional NMT model, reproduced from~\citet{bahdanau2014neural}. English is the source, and French is the target. Lighter colored boxes indicate higher attention scores (black: 0, white: 1).}
    \label{fig:attention}
\end{figure}

First, we turn to \citet{bahdanau2014neural}, who provide us a visualization of alignment weights --- not unlike the word alignment visualization of Figure~\ref{fig:wa_example} --- as reproduced in Figure~\ref{fig:attention}. Inspecting the attention diagram, we see that many of the alignments are indeed quite reasonable.
\citet{bahdanau2014neural} provide as a qualitative example the aligment mapping of source phrase ``the man'' to target phrase ``l' homme''. A word alignment would have instead two alignments ``the'' to ``l''' and ``man'' to ``homme''. They argue that the alignment suggested by attention is better  here because it captures the gender inflection inherent to French.

However in this same diagram, we observe a negative qualitative example. Using attention weights, the target phrase ``said'' maps to the source words ```` , a dit .'' In fact, the magnitude of the weights for ``dit'' and ``.'' are nearly equal, which is intuitively not the case. This is an example of a spurious alignment, indicating that attention captures more than just word alignments. While models like GIZA++ also make spurious alignment, we will see in later sections that their error rate is much lower than this vanilla attentional model. More problematic is the fact that because attention alignments are soft, it is impossible to collect ground-truth labels---we cannot ask humans to perform the imprecise task of assigning real-valued weights between possible aligned words.

\citet{ghader-monz-2017-attention} performs an analysis comparing attention and word alignment, using the attentional model of~\citet{luong-etal-2015-effective}. Their motivating hypothesis is: if attention corresponds to to word alignment, then better consistency between the two should lead to better translations (as \citet{callison-burch-etal-2004-statistical} showed for SMT).
First, they define an \textit{attention loss} between the gold word alignments\footnote{\label{foot:7} The hard word alignments are converted to soft alignments beforehand. The conversion is simple: suppose target word $y_t$ is aligned to the set of source words $A_{y_t}$. Then each alignment has weight $1/|A_{y_t}|.$} and the attention weights.
Next, they define a \textit{word prediction loss}. They find that the correlation between these two losses depends on the part-of-speech tag of the target word. Nouns have a high correlation, whereas verbs have a lower correlation. Despite this, verbs are translated more accurately on average. This shows that when the model translates verbs, it pays attention to more than just the aligned target verb.

\citet{ferrando-costa-jussa-2021-attention-weights} performs a similar analysis, but on a transformer model. They focus on the encoder-decoder attention mechanism and come to a similar conclusion: that attention only \textit{sometimes} correlates with alignments, but it \textit{largely} explains model predictions. They use a previously proposed method~\cite{kobayashi-etal-2020-attention} (which is described in Section~\ref{sec:neural_wa}) for inducing word alignments from a Transformer model. This method makes more alignment errors than GIZA++. They analyze these alignment errors by considering the relative contributions of both parts of the decoder input, at time step $t$: the encoder embeddings (or \textit{source sequence}), and the decoder embeddings of the prior time steps (or \textit{prefix sequence}).
Contributions are calculated through an input perturbation method of either the source or prefix sequence, while keeping the other component unchanged.
This allows them to calculate the proportion of contribution of either sequence.

Intuitively, when generating a target word, if source contribution is high, then the source sequence encodes useful, perhaps alignment-like, information. This happens, for example, when the model is predicting a named entity, and just has to copy it from the source. Otherwise, if the target contribution is high, then the source sequence is less informative.  This happens, for example, when the model is in the middle of generating an idiom such as ``ladies and gentlemen''. In this case, these target tokens are most often incorrectly aligned to \textit{finalizing tokens}, which are either end-of-sentence (\texttt{</s>}) or closing punctuation (\texttt{.}). \citet{ferrando-costa-jussa-2021-attention-weights} conjecture that these wrong alignments occur because the model learns to use these tokens as throwaways, using them as signals to skip attention, and thus skip alignment.

To summarize these findings, in some cases, attention weights correspond well to human intuition on what source and target phrases are aligned. In many other cases they do not, which suggests attention captures other information. To conclude this section, we have shown that attention does not replace the explicit word alignment task.

\section{Neural Approaches toward Word Alignment}
\label{sec:neural_wa}
In this section, we survey the literature on neural word alignment, and classify them into three broad approaches: \textbf{induction}, \textbf{unsupervised}, and \textbf{guided}. We first consider models that operate on the attentional NMT models covered in the prior section. We furthermore describe works that obtain alignments from \textit{multilingual}, encoder-only language models, and also classify them into the three broad approaches.

We focus on those papers that report alignment error rate (AER), the standard metric for evaluating the quality of predicted word alignments respect to gold-standard ones (lower is better). AER is given mathematically as
\begin{equation}
    \text{AER} = 1 - \frac{|A \cap S| + |A \cap P|}{|A| + |S|},
\end{equation}
where $A$ is the set of hypothesized links, and $S$ and $P$ are from a manually annotated set of links\footnote{$S$ure and $P$ossible respectively. For ease of calculation, $P$ is often collapsed into $S$.}

In the main text, we use AER as shorthand for AER on the RWTH German-English (de-en) word alignment test set\footnote{\url{https://www-i6.informatik.rwth-aachen.de/goldAlignment/index.php}} (500 sentence pairs). We cover work that does not report AER for de-en in Appendix~\ref{appsec:nwa}.

Table~\ref{tab:align_aer} summarizes AER results for all aligners covered. Here we report AER for both de-en, and the Hansards English-French (fr-en) test set\footnote{\url{https://web.eecs.umich.edu/~mihalcea/wpt/}} (447 sentence pairs). Unfortunately, the literature is not consistent with reporting statistical system baselines; we therefore use the most recently reported scores for these. 

\begin{table}[ht]
    \centering
    \begin{tabular}{@{}llcc@{}}
        \toprule
        Model & Approach & De-En & Fr-En \\ \midrule
        \citet{luong-etal-2015-effective} & induction & 34.0 & - \\
        \citet{li_word_2019}, 1 & induction & 42.8 & - \\
        \citet{garg-etal-2019-jointly}, 1 & induction & 32.6 & 17.0 \\
        \citet{ding-etal-2019-saliency} & induction & 22.3 & 8.5 \\ 
        \citet{kobayashi-etal-2020-attention} & induction & 25.0 & - \\
        \citet{ferrando-costa-jussa-2021-attention-weights} & induction & 22.1 & - \\
        \citet{chen-etal-2020-accurate}, 1 & induction & \textbf{17.9} & \textbf{6.6} \\
        \midrule
        \citet{zenkel_adding_2019} & unsupervised & 21.2 & 10.0 \\ 
        \citet{zenkel_end--end_2020}, 1 & unsupervised & 17.9 & 8.4 \\
        \citet{chen-etal-2021-mask} & unsupervised & \textbf{14.4} & \textbf{4.4} \\
        \midrule
        \citet{li_word_2019}, 2 & guided & 39.3 & - \\
        \citet{peter2017generating} & guided & 19.0 & - \\
        \citet{garg-etal-2019-jointly}, 2 & guided & 16.0 & \textbf{4.6} \\
        \citet{zenkel_end--end_2020}, 2 & guided & 16.3 & 5.0 \\
        \citet{chen-etal-2020-accurate}, 2 & guided & \textbf{15.8} & 4.7 \\
        \midrule
        \citet{jalili-sabet-etal-2020-simalign} & multilingual (i) & 18.8 & 7.6 \\
        \citet{dou-neubig-2021-word}, 2 & multilingual (u) & 15.0 & 4.1 \\
        \citet{nagata_supervised_2020} & multilingual (g) & 11.4 & 4.0  \\
        \midrule
        Berkeley Aligner & statistical & 20.5 & - \\
        fast\_align & statistical & 27.0 & 10.5 \\ 
        GIZA++ & statistical & 18.7 & 5.5
    \end{tabular}
    \caption{Alignment Error Rates (AER) for several word alignment models, on the German-English and French-English datasets. We report bidirectional (i.e., symmetrized alignment) results where possible, and otherwise unidirectional from source to English.
    The last 2 rows are statistical models, whereas the rest are neural. For papers with multiple models of interest, we append 1 or 2 to the citation. The multilingual models are further subclassified into (i)nduction, (u)nsupervised, and (g)uided.}
    \label{tab:align_aer}
\end{table}

\subsection{Word Alignments through Induction on Attention}
\label{ssec:induct}
In this section, we cover approaches that directly induce word alignments without additional finetuning. We begin by describing an initial hurdle. Word alignments operates on sequences $e, f$ that are translations of each other. However, an attentional NMT model will generate a predicted translation $\hat{f}$ that most likely differs from $f$. We cannot directly compare word alignments between $e, f$ and $e, \hat{f}$. The typical approach is to ``force decode'' the attentional models: at each time step, the gold token $f_i$ is selected, and thus $\hat{f}=f$.

\citet{luong-etal-2015-effective} use force decoding on their global attention model (described in Section~\ref{sssec:att_trans}), and extract alignments by considering, for each target word, the source word with the highest attention weight (we refer to this as the \textit{argmax} approach). They do the same for their local attention models, which looks at a subset of source words at each time step. They find the best model achieves 34 AER.

The following works perform word alignment induction on transformer NMT models. 
\citet{garg-etal-2019-jointly} create a similar, ``naive'' attention baseline by layer-wise averaging of attention probabilities, finding the best AER of 32.6 on layer 5 of a 6-layer transformer.

\citet{kobayashi-etal-2020-attention} propose a simple refinement. Instead of using the attention weight $\alpha$ directly, they used the norm of the weighted projected vector $\lVert \alpha f(x) \rVert$, where $f(x)$ is the transformed input vector.
They modify the argmax approach by selecting the source word $s_j$ that gains the highest weight when \textit{inputting} $t_i$ (instead of when \textit{outputting} $t_i$). This method with norm-based induction achieves 25.0 AER. 

\citet{ferrando-costa-jussa-2021-attention-weights} extend this method by masking out final tokens, and performing a weighted average across attention heads, by a calculated head importance score. Using weight-based induction they achieve 22.1 AER, vs. 18.7 for GIZA++.

\citet{chen-etal-2020-accurate}, similarly to \citet{kobayashi-etal-2020-attention}, induce alignment on a transformer when inputting $t_i$, then average attention weights across heads. Their implementation achieves 17.9 AER, finally beating GIZA++ in 2020.

An alternate line of work uses input perturbation for induction word alignments. \citet{li_word_2019} measure the prediction difference of a target word if a source word is removed, aligning those words with the highest difference. This method achieves 42.8 AER. \citet{ding-etal-2019-saliency} introduce a word saliency method inspired by visual saliency from computer vision. Afterwards they apply smoothing through random sampling. which This approach allows induction of word alignments from any NMT model. Applied to a transformer it achieves 36.4 AER, whil to a convolutional NMT model achieves 27.3 AER.

\subsection{Unsupervised Neural Word Alignments}
\label{ssec:unsuper}
This section discusses \textit{unsupervised} neural approaches to word alignment; i.e., those that do not train on gold word alignments. Our definition of unsupervised means that we exclude those methods that rely on any given word alignments -- even if they come from another system such as GIZA++. In this way, the task of these neural aligners is the same as the statistical ones discussed in Section~\ref{sec:stat_wa}. 

\citet{zenkel_adding_2019} extends the transformer architecture with a separate \textit{alignment layer} on top of the decoder sub-network. This layer differs from the decoder in that there is no self-attention nor skip connections. This essentially forces the alignment layer to rely only on the context vector $c_t$. This model thus predicts target words twice: once for the alignment layer, and once for the decoder layers.
To initialize the attention weights, they perform a forward pass of the transformer, then take the weights from the alignment layer.
This model achieves 26.6 AER. \citet{zenkel_end--end_2020} build on this by incorporating an auxiliary loss function to encourage contiguous attention matrices, achieving 17.9 AER.

\citet{chen-etal-2021-mask} design a self-supervised word alignment model, which in parallel masks out target tokens and recovers it conditioned on the source tokens, and other target tokens. This model uses two variants of attention, which they term \textit{static-KV attention}, which enables masking over target words in parallel, and \textit{leaky attention}, which minimizes incorrect alignments to finalizing tokens. They also perform agreement-based training, and their best model achieves 14.4 AER.

\subsection{Guided Neural Word Alignments}
\label{ssec:super}
This section discusses \textit{guided} neural approaches to word alignment, i.e., those that utilize training data with word alignments. We use ``guided'' instead of ``supervised'' to convey that these models, instead of using gold-annotated alignments, usually are guided silver word alignments generated by GIZA++.

\citet{peter2017generating} extend an attentional NMT model with \textit{target foresight}. In the original attention mechanism, the attention head only knows the prior predicted target words. Target foresight introduces knowledge of the current target word into the calculation of attention. They further train with supervised alignments from GIZA++, achieving a 19.0 AER.

\citet{garg-etal-2019-jointly} propose a single transformer-based model to jointly translate and align. They choose one attention head to supervise with a \textit{guided attention loss}. The best model further considers the entire target context, instead of only the current target word, or prior target words. The self-supervised approach using induced alignments achieves 20.2 AER, whereas supervising with GIZA++ output achieves 16.0 AER. 

\citet{zenkel_end--end_2020} further refine their unsupervised model (see Section~\ref{ssec:unsuper}), which had 17.9 AER, using the guided alignment loss of~\cite{garg-etal-2019-jointly}. The silver alignments come from the first-pass model instead of GIZA++, and after further training, the final model achieves 16.3 AER.

\citet{chen-etal-2020-accurate} similarly use silver alignments from the predictions of their induction-based model (see Section~\ref{ssec:induct}) to train an additional alignment module, achieving 15.8 AER.

\subsection{Word Alignments from Multilingual Language Models}
Transformers were first developed with the machine translation task in mind, and started out as encoder-decoder, bi-lingual language models. However, follow-up work has shown the effectiveness of encoder-only, \textit{multilingual} LMs for various NLP tasks. Examples include mBERT~\cite{devlin-etal-2019-bert} and XLM-roBERTa~\cite{conneau-etal-2020-unsupervised}. Because they are encoder-only, they do not have source-target attention (only self-attention), and  are not trained on any parallel data (only on monolingual datasets in many languages). Still, a line of work has found that high-quality word alignments can be extracted from these models.

\paragraph{Induction} \citet{jalili-sabet-etal-2020-simalign} propose to extract alignments from similarity matrices induced from parallel sentence embeddings. Each cell of this similarity matrix measures some similarity function between source word $x_i$ and target word $y_j$. They first define a method, \textit{Argmax}, to align words $x_i$ and $y_j$ if and only if they are most similar to each other. Then they propose \textit{Itermax}, to iteratively apply \textit{Argmax}, zeroing out similarities at each iteration, until all words are aligned. This method achieves 18.8 AER for our de-en dataset, and beats or is competitive with GIZA++ AER for other language pairs. This suggests that surprisingly, multilingual encoder-only LMs, in the process of training on their masked language modeling tasks, come to an internal notion of word alignments as well.

\paragraph{Unsupervised} \citet{dou-neubig-2021-word} propose to combine the above multilingual LM induction approach with parallel text fine-tuning on two word alignment objectives. The first encourages closer contextualized representations at the aligned word-level, whereas the second does so for representations of parallel sentences. They further introduce a softmax-based alignment between the for the similarity matrices. Their multilingual model achieves 15.0 AER for our de-en task, and similarly high scores for other tasks.

\paragraph{Guided} \citet{nagata_supervised_2020} frame word alignment as a \textit{question-answering} (QA) task, where the context is the target sentence, the question is the source span highlighted in context, and the answer is the aligned target span. They then then train in a standard extractive QA paradigm, achieving 11.4 AER.

\section{Applications of Word Alignment}
\label{sec:app_wa}
Our literature review has shown that researchers in the deep learning era have slowly but steadily been hammering away at the word alignment task. After nearly 2 decades of GIZA++ at the top, researchers have finally surpassed it in the late 2010s. Still, in contrast to other NLP tasks, which have grown rapidly in interest, word alignment has remained fairly niche. Skeptical readers may ask: other than as another leaderboard task, why should I care about the word alignment task today?
 
Broadly speaking, the main application of word alignments is in cross-lingual settings, or settings that go across languages. A prototypical cross-lingual task is, of course, machine translation. The core property that makes word alignments so useful in cross-lingual settings is that alignments can be learned in an unsupervised manner. This was true in the statistical MT era, and remains true today.

In this final section, we broadly survey the literature for applications of word alignment. We identify three main use cases: annotation projection, improving translation, and monolingual word alignments. As applications of the task are agnostic to the modeling approach used (neural, statistical, etc.), we consider works across the decades.

\subsection{Annotation Projection through Word Alignment}
\label{ssec:annot}

\begin{figure}
\centering
    \begin{subfigure}[t]{.47\textwidth}
      \centering
      \includegraphics[width=.75\textwidth]{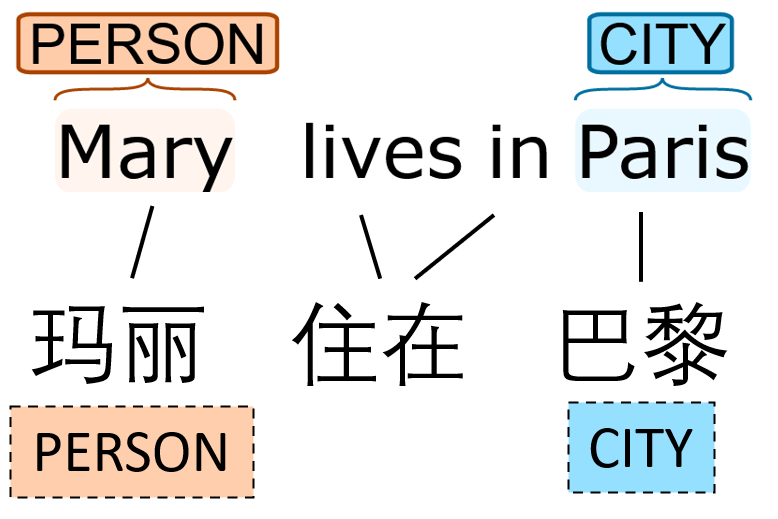}
      \captionof{figure}{The source sentence has NER annotations \texttt{(Mary, PERSON)} and \texttt{(Paris, CITY)}. These are projected to the target sentence, resulting in annotations \texttt{(\begin{CJK*}{UTF8}{gbsn}玛丽\end{CJK*}, PERSON)} and \texttt{(\begin{CJK*}{UTF8}{gbsn}巴黎\end{CJK*}, CITY)}.}
      \label{fig:ner_project}
    \end{subfigure} \qquad %
    \begin{subfigure}[t]{.47\textwidth}
      \centering
      \includegraphics[width=.95\textwidth]{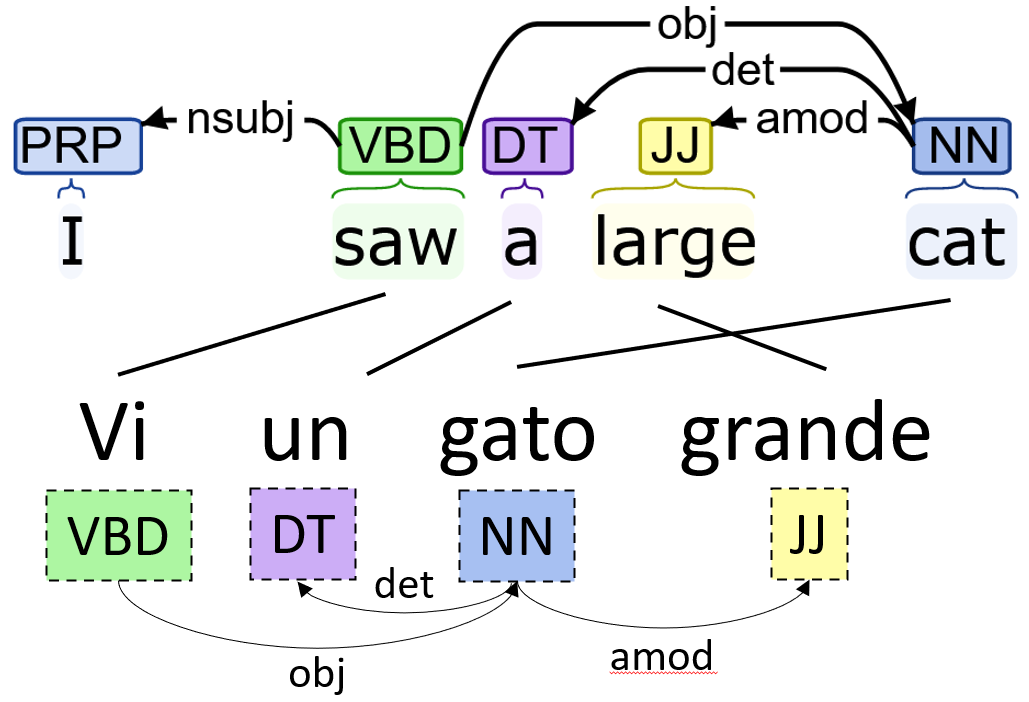}
      \captionof{figure}{First, the part-of-speech annotations are projected. Also projected are the dependency relations represented by the arcs, which relate two words --- a head, and a dependent.}
      \label{fig:dep_project}
    \end{subfigure}
    \caption{Examples of annotation projection for (a) named entity recognition (NER), (b) dependency parsing.}
\end{figure}

Despite advances in unsupervised and self-supervised learning, the performance of supervised learning approaches still dominates for most NLP tasks (as seen in Section~\ref{sec:neural_wa}). Even a minimal amount of supervision goes a long way.
How do we apply our state-of-the-art systems to low-resource languages which have little to no annotated data? Annotation projection is a way to provide supervision in such settings. It has been a major use-case of word alignment historically and to the present.

Annotation projection is the process of transferring annotations on text in a high-resource language (most commonly English) to a low-resource language. It is assumed that a parallel corpus exists, with word-aligned parallel sentences. In this way, we can first train a model using available annotated data for some task in the source language. We use this model to generate annotations on the source side of the parallel bitext. Then we use the word alignments to project source labels to target labels. We now have labeled examples in the target language, which we can use to train a new model on the task. Example\footnote{The top half of these diagrams are generated using \url{https://corenlp.run/}.} annotation projections are shown in Figure~\ref{fig:ner_project} and~\ref{fig:dep_project}.

\paragraph{Formalizing Word Alignment}
Annotation projection addresses the following setting (with reference to ~\citet{rasooliCrossLingualTransferNatural2019}).
Suppose we want a system which can solve a task for some given sentences $\{t^{(i)}\}^m_{i=1}$ in a low-resource language $L_t$. For each sentence we have source language sentences $\{s^{(i)}\}^m_{i=1}$, where $s^{(i)}$ and $t^{(i)}$ are translations for all $i$. Furthermore, we have word alignments between these sentences.

We first have supervised model $\mathcal{M}_{sup}$ learn a task from labeled source language examples $\mathcal{D}=\{(x_j, y_j)\}^n_{j=1}$. We apply the trained model to the source-side translated text $\{s_i\}^m_{i=1}$ to obtain predictions $\{y_{s_i}\}^m_{i=1}$. Now, we project each source prediction to obtain the target labels $y_{t_i}$. 

The annotation projection process gives us labeled target language examples $\mathcal{D}_t=\{(t^{(i)}, y_{t_i})\}^m_{i=1}$, from which we can train a model $\mathcal{M}_{proj}$. We can now use system $\mathcal{M}_{proj}$ to solve the task in the target language.

The key insight is that reasonable word alignments can be learned in an unsupervised manner. Acquiring human-labeled data for a specific NLP task in a low-resource language can be difficult and expensive. Acquiring translations from this low-resource language to a high-resource one is less so, and after running a unsupervised word alignment system over these translations, we can use annotation projection to create models for any NLP task in the low-resource language. The projected annotations are imperfect, of course, given both a) errors in the projection processes, such as incorrect word alignments, and b) differences in how languages represent (or do not represent) meaning and features. But the technique remains a reasonable, effective start for addressing various tasks and domains in low-resource languages or domains.

\subsubsection{Annotation Projection for NLP Tasks}
\citet{hwa2005bootstrapping} apply annotation projection to parsing, the task of predicting a tree relating the words of a sentence. They first directly project English trees to Spanish and Chinese, finding the projected trees are quite lacking. They then perform post-projection transformation of these trees based on linguistic knowledge of the target languages. Finally, they train target language syntactic parsers on these transformed trees, finding the process allows for reasonable bootstrapping of syntactic parsers for these languages. Follow-up work improves syntactic parsing performance by incorporating posterior regularization~\cite{ganchev-etal-2009-dependency}, by using parallel guidance and entropy regularization~\cite{ma-xia-2014-unsupervised}, and by iteratively learning over from more dense to less dense dependency structures~\cite{rasooli-collins-2015-density, rasooli-collins-2017-cross}.

The annotation projection process itself is generally kept constant; research instead involves changing the model, the representation of the task, or the refinements to the projection models. As those do not involve word alignments, and are thus beyond the scope of the current tutorial, we omit further details. We refer interested readers to several papers by Mohammad Sadegh Rasooli, Maryam Aminian, and collaborators: annotation projection for semantic role labeling~\cite{aminian-etal-2017-transferring,aminian-etal-2019-cross}, for sentiment analysis~\cite{rasooli2018cross}, and for broad-coverage semantic dependency parsing~\cite{aminian-etal-2020-multitask}. We discuss non-NLP use cases of annotation projection, and word alignment in general, in the following sections.

\subsubsection{Annotation Projection for Medical Terms} 
\citet{TranslatingMedicalTerminologies2009} uses annotation projection for the medical domain.  They note that the increasing internationalization of medicine necessitates translation of medical terminology. Because of the specific, technical nature of the domain, this is difficult even for human translators. Therefore, they propose  a human-in-the-loop system in which humans are given candidate aligned medical terms, which they can then inspect, then correct and filter out if needed.

We highlight the major steps of their proposed system here. First, they collect a parallel corpus, using a Canadian health website which has documents in both English and French. Then, they align sentence translations within these relatively parallel documents using an existing system. They perform automatic word alignment on these sentence pairs, then refine the alignments using human annotators. Finally, they aggregate similar alignments together into a relational database, and have human annotators select and refine them, arriving at a final set of translated medical terminologies.

They find their methodology creates English-French medical dictionaries with reasonably high-precision entries. From 50k sentence pairs, they obtain 15k filtered alignments. A manual review of 2127 terms finds a 79\% precision (vs. 41\% before filtering). \citet{nystrom2006creating} use the same a similar method for the English-Swedish pair, but use aligned terms instead of aligned sentences, resulting in more limited coverage.

\subsection{Word Alignments for Improving Neural Machine Translation}
\label{ssec:super_wa}
Let us now return to machine translation, the task which initially motivated word alignments. For statistical MT, word alignment allows for decomposition of translating an entire sentence to translating parts of a sentence. But what about for neural MT? We have seen that the attention mechanism, at the core of modern day transformer models, captures some notion of alignment. We have also seen that high-quality alignments can be induced from both encoder-decoder models and encoder-only models. Still, follow-up work has shown that word alignment can additionally inform neural MT models. 

A major issue with NMT systems is that they are prone to mistranslating low-frequency words, especially without proper sampling algorithms. This is a side-effect of their treating vocabulary words as embeddings, or real-valued vectors. This representation allows embedding notions of similarity between related words. However, a resulting drawback is that NMT systems are more likely to mistranslate words that make sense in a given context, but actually do not reflect the source sentence. It is often the case that low-frequency words are often the main content words, so effort should be made to properly translate them. In contrast, SMT systems avoid this problem due to their usage of explicit, discrete translation tables.

\citet{arthurIncorporatingDiscreteTranslation2016} give an example for English-to-Japanese translation, in which the system mistakenly translates the word for ``Tunisia'' into the word for ``Norway''. This is likely because ``Norway'' was seen more often in the training distribution. With this motivating example, \citet{arthurIncorporatingDiscreteTranslation2016} address this by incorporating an external, discrete lexicon---collected through word alignments---into NMT models. They first transform the translation table probabilities into a next-word prediction probability, then incorporate this probability into NMT. We now summarize this incorporation process. Given an input sentence $F$, they construct a matrix $L_F$ of shape $|V_e| \times |F|$, where $V_e$ is the target vocabulary. Each cell then specifies $Pr(e_i=n| f_j)$; these probabilities come from the t-tables learned by word alignment. They then repurpose the attention vector $\alpha_t$ to weight each column of $L_f$, such that
\begin{align}
    Pr_{\text{lex}}(e_i|F, e_{1}^{i-1}) = L_F \alpha_t ,
\end{align}
where $e_{1}^{i-1}=e_1,...,e_{i-1}$.

$\alpha_t$ is computed in much the same way, differing in that it is over the target vocabulary, instead of the source tokens. They now need to incorporate this next-word probability with that of the NMT model. They find the best overall method is by using it as a bias within the NMT model's probability distribution,
\begin{align}
    Pr_{\text{bias}}(e_i|F, e_{1}^{i-1}) = \softmax(W_s \eta_i + b_s + \log{ Pr_{\text{lex}}(e_i|F, e_{1}^{i-1}) + \epsilon})) ,
\end{align}
where $W_s \eta_i + b_s$ is from the original NMT model equation, and $\epsilon$ is a hyper-parameter that biases towards using the lexicon probabilities when smaller.

\citet{arthurIncorporatingDiscreteTranslation2016} compare a baseline attentional NMT model to one that incorporates the modified probability. On a English-to-Japanese translation task, they find increases for both translation accuracy (23.2 vs. 20.9 BLEU\footnote{BLEU~\cite{papineni-etal-2002-bleu} is an evaluation metric for MT, based on modified n-gram precision.}), recall of low-frequency words (19.3 vs. 17.7). They further find that their proposed model achieves higher BLEU scores in earlier training steps, suggesting the model can bootstrap learning from the lexical information provided by word alignments. \citet{mi-etal-2016-vocabulary} take a similar, but simpler approach, leveraging word alignment information to limit the vocabulary size predicted by each sentence.

\citet{cohn-etal-2016-incorporating} likewise seek to incorporate information from the word alignment process into attentional NMT. Instead of using word alignments directly, they take the structural biases used in prior statistical, unsupervised word alignment, and incorporate them into NMT models.
First, they incorporate absolute position from IBM Model 2, which adds in a bias towards aligning source and target words at similar relative positions within their sentences.
Second, they incorporate the Markov condition from HMM aligners~\cite{vogel-etal-1996-hmm}, which captures locality in the source language. This means the assumption that if two target words are contiguous, it is likely that their aligned words are contiguous.
Third, they model fertility, which captures the number of aligned source words to a target word.
Fourth, inspired by~\citet{liang-etal-2006-alignment} they jointly train two models in parallel, one in each direction. On several language pairs in low-resource settings, they find their model trained with these biases achieves BLEU increases over a baseline attentional NMT model. For example Chinese-English BLEU is 44.1 vs. 41.2.

\citet{raganatoEmpiricalInvestigationWord2021} proposes to supervise existing zero-shot\footnote{Zero-shot means to perform the task without having seen relevant training data -- here the language pairs of interest.} multilingual NMT systems with word alignment information. They follow the joint training of translation and alignment from~\cite{garg-etal-2019-jointly}, and produce word alignments using~\cite{dou-neubig-2021-word}. They compare their model to a transformer baseline, and find that while average BLEU scores are the same across seen language pairs en-xx and xx-en (where xx is a non-English language), the word alignment supervised models achieve a 1.9 BLEU increase for unseen pairs (11.9 vs. 10.0).

We next consider the dictionary-guided MT task, where the input includes a dictionary of suggested translations. This is used in the case of technical domains, such as medical or information technology, in which users would like to enforce translations of given domain-specific terms. \citet{chatterjeeGuidingNeuralMachine2017} extend a NMT decoder with a guidance mechanism, which uses supervision from learned word alignments, to constrain decoding. This increases en-de BLEU (25.5 vs 21.7). \citet{alkhouli-etal-2018-alignment} augment a transformer NMT model with an alignment head, which models its source context as a binary-valued vector; 1 if there is an alignment, 0 otherwise. This is concatenated to the attention head, and the model increases en-ro BLEU (31.0 vs. 29.7). \citet{songAlignmentEnhancedTransformerConstraining2020} instead use a dedicated attention head to emulate the external alignments, and calculate a separate attention loss, finding similar improvements.

\subsection{Word alignment for Computer-Assisted Translation (CAT)}
A related field to machine translation is computer-assisted translation (CAT). Instead of using systems to directly translate a source text, CAT instead develops systems to assist human translators. CAT involves translation for translators (who are necessarily bilingual), whereas MT involves translation for more general end-users (who are likely monolingual). CAT is commonly used when translating technical documents, which contain HTML tags and other markup. These tags are not natural language, and must be handled separately by any MT and CAT systems. An example of markup is that if a source term is bolded, the corresponding target term should remain bolded after translation. Many authors have explored using annotation projection as part of the CAT processing pipeline, to suggest possible tags for translators~\cite{tezcan-vandeghinste-2011-smt,joanis-etal-2013-transferring,arcan2017leveraging,mullerTreatmentMarkupStatistical2017}.

Other work has used word alignments to improve the translation memory (TM) component of CAT systems. TMs are databases that store previously translated segments (sentences, paragraphs, etc.) that human translators can refer to; they are used to assist in nearly all CAT tasks. ~\citet{wu-etal-2005-improving} propose to enhance TMs using word alignment information at several level of matching. For the sentence-level and sub-sentential level matching metrics, they incorporate into each an alignment confidence score between source and target sentence. For pattern-based machine translation, they parse the source sentences into phrases, then use the word alignment phrase tables to extract suggested translations. They find their system improves translation quality and saves 20\% translation time for humans.

\citet{koehn-senellart-2010-convergence} integrate concepts from both TM and SMT systems to improve CAT systems. The proposed method first finds a fuzzy-matched TM sentence to the source sentence, then identifies the differing words between the two (they do the latter by edit distance). They use the target sentence from the TM for the common words, and use an SMT system to translate the differing words. Finally, they use word alignments on the TM example to project from the source to the target span, and replace that span with the SMT translation.  Their results show that when the fuzzy match scores are $\geq 80\%$, the combined SMT+TM approach achieves higher BLEU performance than either alone.

Moving out of CAT, \citet{heFastAccurateNeural2021} propose a model which incorporates translation memories into NMT. They extend a transformer model with a TM component of a single sentence pair. They consider three ways of encoding the TM, the best of which weights the sentence similarity scores by an external word alignment (obtained from fast\_align), so that the model pays more attention to aligned words when considering the TM. Compared to NMT systems which have larger TMs, i.e., they retrieve multiple sentences, their system is faster and more accurate, showing that using word alignment information assists TMs used for NMT as well as CAT.

\subsection{Other use-cases for word alignment}
Finally, we discuss other use cases for word alignment.

\subsubsection{Monolingual Word Alignment}
Thus far we have considered only the word alignment task in cross-lingual settings. But just  as translation is one instance of a sequence-to-sequence task, so too cross-lingual word alignment is one instance of word alignment. \textit{Monolingual word alignment} is the task of aligning words between two related sentences in the same language.

Early work on monolingual word alignment addresses natural language inference (NLI)~\cite{maccartneyPhraseBasedAlignmentModel2008}, the task of determining if a natural language hypothesis $H$ can be entailed by a given premise $P$. Despite the difference between the MT and NLI tasks\footnote{Other than the obvious monolinguality, NLI also differs in the word length asymmetry, given $P$ is usually much longer than $H$}, they train GIZA++ and other supervised (MT) word aligners on NLI sentence pairs, achieving 74.1\% F1 on a human word-aligned NLI test set. Their proposed system, however, which more effectively leverages aspects of the NLI problem by techniques such as using external semantic relatedness information, and a phrase-based representation of alignment, achieves 85.5\% F1.

Follow-up work improves this model by using an integer linear programming (ILP) based exact decoding technique~\cite{thadani-mckeown-2011-optimal}. Other work extends monolingual alignment to paraphrase alignment~\cite{thadani-etal-2012-joint}, question answering~\cite{yao2014feature}, and semantic textual similarity~\cite{li-srikumar-2016-exploiting}.

More recently, ~\citet{lan-etal-2021-neural} introduce a human-annotated benchmark for monolingual word alignment. They further propose a neural semi-Markov CRF alignment model, which unifies word-level and phrase-level alignments, captures semantic similarity between source and target spans, and incorporates Markov transition probabilities. Their proposed model outperforms prior models on their dataset, as well as on three prior datasets. They further show that monolingual word alignments have useful downstream applications for text simplification tasks, building on~\citet{jiang-etal-2020-neural}, and for several sentence pair classification tasks, following~\citet{lan-xu-2018-neural}.

\subsubsection{Improving MT Evaluation}
We now discuss a specific use case of monolingual word alignment: in improving metrics for machine translation evaluation.

We have seen the standard automated evaluation metric for word alignment is AER, and briefly touched on a metric for machine translation, BLEU~\cite{papineni-etal-2002-bleu}. BLEU is a precision-based metric that is easy to calculate and to understand. However, the simplicity of BLEU leads to several issues. Perhaps most notably is that it entirely fails to capture semantic similarity. For example, suppose the MT system generates ``a big dog'', and the gold reference is ``one large canine''. BLEU gives a score of 0, even though the phrases are essentially semantically equivalent.

METEOR~\cite{lavie-agarwal-2007-meteor} is an automated evaluation metric for evaluating machine translation that addresses these shortcomings of BLEU. It performs scoring by aligning hypotheses to one or more reference translations. A monolingual word aligner aligns words and phrases in successive stages: exact match, by stem, by synonym, and by paraphrase. METEOR then scores the computed alignment using unigram precision, recall, and a measure of explicit ordering. METEOR correlates well with human judgment at the sentence-level, and is another commonly-used metric for evaluating machine translation.


\section{Conclusion}
\label{sec:concl}
Let us conclude by recapping what we have covered in this tutorial. We began by formalizing the tasks of word alignment and statistical machine translation, finding that they are intrinsically linked. We the turned to neural machine translation, taking a detour to describe a basic encoder-decoder NMT model. We got back on course by formalizing the attention mechanism, which introduces a concept of alignment between tokens in the source and target sentences. We found that attention captures some notion of alignment, but also other linguistic information. We then moved to a survey approach. We performed a comprehensive literature review of neural word aligners, finding that the task remains a niche, yet underexplored task in the deep learning era. Finally, we surveyed the applications of word alignment, from annotation projection to improving translation.

We hope that our tutorial has instilled in readers not only a past and present understanding of the word alignment task, but also an interest to future word alignment research ahead.

\begin{acknowledgments}
This report was written in partial fulfillment of the WPE-II requirement for the PhD in Computer and Information Science at the University of Pennsylvania. The author would like to thank Professors Mark Liberman, Wei Xu, Chris Callison-Burch and Benjamin Pierce, for their guidance throughout the process. The author would also like to thank Veronica Qing Lyu, Weiqiu You, and Harry Li Zhang for their peer reviews.
\end{acknowledgments}

\bibliography{main, anthology}

\begin{thebibliography}{73}
\expandafter\ifx\csname natexlab\endcsname\relax\def\natexlab#1{#1}\fi

\bibitem[{Alkhouli, Bretschner, and Ney(2018)}]{alkhouli-etal-2018-alignment}
Alkhouli, Tamer, Gabriel Bretschner, and Hermann Ney. 2018.
\newblock On the alignment problem in multi-head attention-based neural machine
  translation.
\newblock In \emph{Proceedings of the Third Conference on Machine Translation:
  Research Papers}, pages 177--185, Association for Computational Linguistics,
  Brussels, Belgium.

\bibitem[{Aminian, Rasooli, and Diab(2017)}]{aminian-etal-2017-transferring}
Aminian, Maryam, Mohammad~Sadegh Rasooli, and Mona Diab. 2017.
\newblock Transferring semantic roles using translation and syntactic
  information.
\newblock In \emph{Proceedings of the Eighth International Joint Conference on
  Natural Language Processing (Volume 2: Short Papers)}, pages 13--19, Asian
  Federation of Natural Language Processing, Taipei, Taiwan.

\bibitem[{Aminian, Rasooli, and Diab(2019)}]{aminian-etal-2019-cross}
Aminian, Maryam, Mohammad~Sadegh Rasooli, and Mona Diab. 2019.
\newblock Cross-lingual transfer of semantic roles: From raw text to semantic
  roles.
\newblock In \emph{Proceedings of the 13th International Conference on
  Computational Semantics - Long Papers}, pages 200--210, Association for
  Computational Linguistics, Gothenburg, Sweden.

\bibitem[{Aminian, Rasooli, and Diab(2020)}]{aminian-etal-2020-multitask}
Aminian, Maryam, Mohammad~Sadegh Rasooli, and Mona Diab. 2020.
\newblock Multitask learning for cross-lingual transfer of broad-coverage
  semantic dependencies.
\newblock In \emph{Proceedings of the 2020 Conference on Empirical Methods in
  Natural Language Processing (EMNLP)}, pages 8268--8274, Association for
  Computational Linguistics, Online.

\bibitem[{Arcan et~al.(2017)Arcan, Turchi, Tonelli, and
  Buitelaar}]{arcan2017leveraging}
Arcan, Mihael, Marco Turchi, Sara Tonelli, and Paul Buitelaar. 2017.
\newblock Leveraging bilingual terminology to improve machine translation in a
  cat environment.
\newblock \emph{Natural Language Engineering}, 23(5):763--788.

\bibitem[{Arthur, Neubig, and
  Nakamura(2016)}]{arthurIncorporatingDiscreteTranslation2016}
Arthur, Philip, Graham Neubig, and Satoshi Nakamura. 2016.
\newblock Incorporating {{Discrete Translation Lexicons}} into {{Neural Machine
  Translation}}.
\newblock In \emph{Proceedings of the 2016 {{Conference}} on {{Empirical
  Methods}} in {{Natural}} {{Language Processing}}}, pages 1557--1567,
  {Association for Computational Linguistics}, {Austin, Texas}.

\bibitem[{Bahdanau, Cho, and Bengio(2015)}]{bahdanau2014neural}
Bahdanau, Dzmitry, Kyunghyun Cho, and Yoshua Bengio. 2015.
\newblock Neural machine translation by jointly learning to align and
  translate.
\newblock In \emph{3rd International Conference on Learning Representations,
  {ICLR} 2015, San Diego, CA, USA, May 7-9, 2015, Conference Track
  Proceedings}.

\bibitem[{Brown et~al.(1993)Brown, Della~Pietra, Della~Pietra, and
  Mercer}]{brown1993mathematics}
Brown, Peter~F, Stephen~A Della~Pietra, Vincent~J Della~Pietra, and Robert~L
  Mercer. 1993.
\newblock The mathematics of statistical machine translation: Parameter
  estimation.
\newblock \emph{Computational linguistics}, 19(2):263--311.

\bibitem[{Callison-Burch, Talbot, and
  Osborne(2004)}]{callison-burch-etal-2004-statistical}
Callison-Burch, Chris, David Talbot, and Miles Osborne. 2004.
\newblock Statistical machine translation with word- and sentence-aligned
  parallel corpora.
\newblock In \emph{Proceedings of the 42nd Annual Meeting of the Association
  for Computational Linguistics ({ACL}-04)}, pages 175--182, Barcelona, Spain.

\bibitem[{Chatterjee et~al.(2017)Chatterjee, Negri, Turchi, Federico, Specia,
  and Blain}]{chatterjeeGuidingNeuralMachine2017}
Chatterjee, Rajen, Matteo Negri, Marco Turchi, Marcello Federico, Lucia Specia,
  and Fr{\'e}d{\'e}ric Blain. 2017.
\newblock Guiding {{Neural Machine Translation Decoding}} with {{External
  Knowledge}}.
\newblock In \emph{Proceedings of the {{Second Conference}} on {{Machine
  Translation}}}, pages 157--168, {Association for Computational Linguistics},
  {Copenhagen, Denmark}.

\bibitem[{Chen, Sun, and Liu(2021)}]{chen-etal-2021-mask}
Chen, Chi, Maosong Sun, and Yang Liu. 2021.
\newblock Mask-align: Self-supervised neural word alignment.
\newblock In \emph{Proceedings of the 59th Annual Meeting of the Association
  for Computational Linguistics and the 11th International Joint Conference on
  Natural Language Processing (Volume 1: Long Papers)}, pages 4781--4791,
  Association for Computational Linguistics, Online.

\bibitem[{Chen et~al.(2020)Chen, Liu, Chen, Jiang, and
  Liu}]{chen-etal-2020-accurate}
Chen, Yun, Yang Liu, Guanhua Chen, Xin Jiang, and Qun Liu. 2020.
\newblock Accurate word alignment induction from neural machine translation.
\newblock In \emph{Proceedings of the 2020 Conference on Empirical Methods in
  Natural Language Processing (EMNLP)}, pages 566--576, Association for
  Computational Linguistics, Online.

\bibitem[{Cheng et~al.(2016)Cheng, Shen, He, He, Wu, Sun, and
  Liu}]{cheng2016agreement}
Cheng, Yong, Shiqi Shen, Zhongjun He, Wei He, Hua Wu, Maosong Sun, and Yang
  Liu. 2016.
\newblock Agreement-based joint training for bidirectional attention-based
  neural machine translation.
\newblock In \emph{Proceedings of the Twenty-Fifth International Joint
  Conference on Artificial Intelligence}, pages 2761--2767.

\bibitem[{Cohn et~al.(2016)Cohn, Hoang, Vymolova, Yao, Dyer, and
  Haffari}]{cohn-etal-2016-incorporating}
Cohn, Trevor, Cong Duy~Vu Hoang, Ekaterina Vymolova, Kaisheng Yao, Chris Dyer,
  and Gholamreza Haffari. 2016.
\newblock Incorporating structural alignment biases into an attentional neural
  translation model.
\newblock In \emph{Proceedings of the 2016 Conference of the North {A}merican
  Chapter of the Association for Computational Linguistics: Human Language
  Technologies}, pages 876--885, Association for Computational Linguistics, San
  Diego, California.

\bibitem[{Conneau et~al.(2020)Conneau, Khandelwal, Goyal, Chaudhary, Wenzek,
  Guzm{\'a}n, Grave, Ott, Zettlemoyer, and
  Stoyanov}]{conneau-etal-2020-unsupervised}
Conneau, Alexis, Kartikay Khandelwal, Naman Goyal, Vishrav Chaudhary, Guillaume
  Wenzek, Francisco Guzm{\'a}n, Edouard Grave, Myle Ott, Luke Zettlemoyer, and
  Veselin Stoyanov. 2020.
\newblock Unsupervised cross-lingual representation learning at scale.
\newblock In \emph{Proceedings of the 58th Annual Meeting of the Association
  for Computational Linguistics}, pages 8440--8451, Association for
  Computational Linguistics, Online.

\bibitem[{Del{\'e}ger, Merkel, and
  Zweigenbaum(2009)}]{TranslatingMedicalTerminologies2009}
Del{\'e}ger, Louise, Magnus Merkel, and Pierre Zweigenbaum. 2009.
\newblock Translating medical terminologies through word alignment in parallel
  text corpora.
\newblock \emph{Journal of Biomedical Informatics}, 42(4).

\bibitem[{Devlin et~al.(2019)Devlin, Chang, Lee, and
  Toutanova}]{devlin-etal-2019-bert}
Devlin, Jacob, Ming-Wei Chang, Kenton Lee, and Kristina Toutanova. 2019.
\newblock {BERT}: Pre-training of deep bidirectional transformers for language
  understanding.
\newblock In \emph{Proceedings of the 2019 Conference of the North {A}merican
  Chapter of the Association for Computational Linguistics: Human Language
  Technologies, Volume 1 (Long and Short Papers)}, pages 4171--4186,
  Association for Computational Linguistics, Minneapolis, Minnesota.

\bibitem[{Ding, Xu, and Koehn(2019)}]{ding-etal-2019-saliency}
Ding, Shuoyang, Hainan Xu, and Philipp Koehn. 2019.
\newblock Saliency-driven word alignment interpretation for neural machine
  translation.
\newblock In \emph{Proceedings of the Fourth Conference on Machine Translation
  (Volume 1: Research Papers)}, pages 1--12, Association for Computational
  Linguistics, Florence, Italy.

\bibitem[{Dou and Neubig(2021)}]{dou-neubig-2021-word}
Dou, Zi-Yi and Graham Neubig. 2021.
\newblock Word alignment by fine-tuning embeddings on parallel corpora.
\newblock In \emph{Proceedings of the 16th Conference of the European Chapter
  of the Association for Computational Linguistics: Main Volume}, pages
  2112--2128, Association for Computational Linguistics, Online.

\bibitem[{Dyer, Chahuneau, and Smith(2013)}]{dyer-etal-2013-simple}
Dyer, Chris, Victor Chahuneau, and Noah~A. Smith. 2013.
\newblock A simple, fast, and effective reparameterization of {IBM} model 2.
\newblock In \emph{Proceedings of the 2013 Conference of the North {A}merican
  Chapter of the Association for Computational Linguistics: Human Language
  Technologies}, pages 644--648, Association for Computational Linguistics,
  Atlanta, Georgia.

\bibitem[{Ferrando and
  Costa-juss{\`a}(2021)}]{ferrando-costa-jussa-2021-attention-weights}
Ferrando, Javier and Marta~R. Costa-juss{\`a}. 2021.
\newblock Attention weights in transformer {NMT} fail aligning words between
  sequences but largely explain model predictions.
\newblock In \emph{Findings of the Association for Computational Linguistics:
  EMNLP 2021}, pages 434--443, Association for Computational Linguistics, Punta
  Cana, Dominican Republic.

\bibitem[{Ganchev, Gillenwater, and
  Taskar(2009)}]{ganchev-etal-2009-dependency}
Ganchev, Kuzman, Jennifer Gillenwater, and Ben Taskar. 2009.
\newblock Dependency grammar induction via bitext projection constraints.
\newblock In \emph{Proceedings of the Joint Conference of the 47th Annual
  Meeting of the {ACL} and the 4th International Joint Conference on Natural
  Language Processing of the {AFNLP}}, pages 369--377, Association for
  Computational Linguistics, Suntec, Singapore.

\bibitem[{Garg et~al.(2019)Garg, Peitz, Nallasamy, and
  Paulik}]{garg-etal-2019-jointly}
Garg, Sarthak, Stephan Peitz, Udhyakumar Nallasamy, and Matthias Paulik. 2019.
\newblock Jointly learning to align and translate with transformer models.
\newblock In \emph{Proceedings of the 2019 Conference on Empirical Methods in
  Natural Language Processing and the 9th International Joint Conference on
  Natural Language Processing (EMNLP-IJCNLP)}, pages 4453--4462, Association
  for Computational Linguistics, Hong Kong, China.

\bibitem[{Ghader and Monz(2017)}]{ghader-monz-2017-attention}
Ghader, Hamidreza and Christof Monz. 2017.
\newblock What does attention in neural machine translation pay attention to?
\newblock In \emph{Proceedings of the Eighth International Joint Conference on
  Natural Language Processing (Volume 1: Long Papers)}, pages 30--39, Asian
  Federation of Natural Language Processing, Taipei, Taiwan.

\bibitem[{He et~al.(2021)He, Huang, Cui, Li, and
  Liu}]{heFastAccurateNeural2021}
He, Qiuxiang, Guoping Huang, Qu~Cui, Li~Li, and Lemao Liu. 2021.
\newblock Fast and {{Accurate Neural Machine Translation}} with {{Translation
  Memory}}.
\newblock In \emph{Proceedings of the 59th {{Annual Meeting}} of the
  {{Association}} for {{Computational Linguistics}} and the 11th
  {{International Joint Conference}} on {{Natural Language Processing}}
  ({{Volume}} 1: {{Long Papers}})}, pages 3170--3180, {Association for
  Computational Linguistics}, {Online}.

\bibitem[{Heafield(2011)}]{heafield-2011-kenlm}
Heafield, Kenneth. 2011.
\newblock {K}en{LM}: Faster and smaller language model queries.
\newblock In \emph{Proceedings of the Sixth Workshop on Statistical Machine
  Translation}, pages 187--197, Association for Computational Linguistics,
  Edinburgh, Scotland.

\bibitem[{Hwa et~al.(2005)Hwa, Resnik, Weinberg, Cabezas, and
  Kolak}]{hwa2005bootstrapping}
Hwa, Rebecca, Philip Resnik, Amy Weinberg, Clara Cabezas, and Okan Kolak. 2005.
\newblock Bootstrapping parsers via syntactic projection across parallel texts.
\newblock \emph{Natural language engineering}, 11(3):311--325.

\bibitem[{Ippolito(2022)}]{ippolito2022nlg}
Ippolito, Daphne. 2022.
\newblock A tutorial on neural language models and text generation.

\bibitem[{Jalili~Sabet et~al.(2020)Jalili~Sabet, Dufter, Yvon, and
  Sch{\"u}tze}]{jalili-sabet-etal-2020-simalign}
Jalili~Sabet, Masoud, Philipp Dufter, Fran{\c{c}}ois Yvon, and Hinrich
  Sch{\"u}tze. 2020.
\newblock {S}im{A}lign: High quality word alignments without parallel training
  data using static and contextualized embeddings.
\newblock In \emph{Findings of the Association for Computational Linguistics:
  EMNLP 2020}, pages 1627--1643, Association for Computational Linguistics,
  Online.

\bibitem[{Jiang et~al.(2020)Jiang, Maddela, Lan, Zhong, and
  Xu}]{jiang-etal-2020-neural}
Jiang, Chao, Mounica Maddela, Wuwei Lan, Yang Zhong, and Wei Xu. 2020.
\newblock Neural {CRF} model for sentence alignment in text simplification.
\newblock In \emph{Proceedings of the 58th Annual Meeting of the Association
  for Computational Linguistics}, pages 7943--7960, Association for
  Computational Linguistics, Online.

\bibitem[{Joanis et~al.(2013)Joanis, Stewart, Larkin, and
  Kuhn}]{joanis-etal-2013-transferring}
Joanis, Eric, Darlene Stewart, Samuel Larkin, and Roland Kuhn. 2013.
\newblock Transferring markup tags in statistical machine translation: a
  two-stream approach.
\newblock In \emph{Proceedings of the 2nd Workshop on Post-editing Technology
  and Practice}, Nice, France.

\bibitem[{Kobayashi et~al.(2020)Kobayashi, Kuribayashi, Yokoi, and
  Inui}]{kobayashi-etal-2020-attention}
Kobayashi, Goro, Tatsuki Kuribayashi, Sho Yokoi, and Kentaro Inui. 2020.
\newblock Attention is not only a weight: Analyzing transformers with vector
  norms.
\newblock In \emph{Proceedings of the 2020 Conference on Empirical Methods in
  Natural Language Processing (EMNLP)}, pages 7057--7075, Association for
  Computational Linguistics, Online.

\bibitem[{Koehn et~al.(2007)Koehn, Hoang, Birch, Callison-Burch, Federico,
  Bertoldi, Cowan, Shen, Moran, Zens, Dyer, Bojar, Constantin, and
  Herbst}]{koehn-etal-2007-moses}
Koehn, Philipp, Hieu Hoang, Alexandra Birch, Chris Callison-Burch, Marcello
  Federico, Nicola Bertoldi, Brooke Cowan, Wade Shen, Christine Moran, Richard
  Zens, Chris Dyer, Ond{\v{r}}ej Bojar, Alexandra Constantin, and Evan Herbst.
  2007.
\newblock {M}oses: Open source toolkit for statistical machine translation.
\newblock In \emph{Proceedings of the 45th Annual Meeting of the Association
  for Computational Linguistics Companion Volume Proceedings of the Demo and
  Poster Sessions}, pages 177--180, Association for Computational Linguistics,
  Prague, Czech Republic.

\bibitem[{Koehn and Senellart(2010)}]{koehn-senellart-2010-convergence}
Koehn, Philipp and Jean Senellart. 2010.
\newblock Convergence of translation memory and statistical machine
  translation.
\newblock In \emph{Proceedings of the Second Joint EM+/CNGL Workshop: Bringing
  MT to the User: Research on Integrating MT in the Translation Industry},
  pages 21--32, Association for Machine Translation in the Americas, Denver,
  Colorado, USA.

\bibitem[{Lan, Jiang, and Xu(2021)}]{lan-etal-2021-neural}
Lan, Wuwei, Chao Jiang, and Wei Xu. 2021.
\newblock Neural semi-{M}arkov {CRF} for monolingual word alignment.
\newblock In \emph{Proceedings of the 59th Annual Meeting of the Association
  for Computational Linguistics and the 11th International Joint Conference on
  Natural Language Processing (Volume 1: Long Papers)}, pages 6815--6828,
  Association for Computational Linguistics, Online.

\bibitem[{Lan and Xu(2018)}]{lan-xu-2018-neural}
Lan, Wuwei and Wei Xu. 2018.
\newblock Neural network models for paraphrase identification, semantic textual
  similarity, natural language inference, and question answering.
\newblock In \emph{Proceedings of the 27th International Conference on
  Computational Linguistics}, pages 3890--3902, Association for Computational
  Linguistics, Santa Fe, New Mexico, USA.

\bibitem[{Lavie and Agarwal(2007)}]{lavie-agarwal-2007-meteor}
Lavie, Alon and Abhaya Agarwal. 2007.
\newblock {METEOR}: An automatic metric for {MT} evaluation with high levels of
  correlation with human judgments.
\newblock In \emph{Proceedings of the Second Workshop on Statistical Machine
  Translation}, pages 228--231, Association for Computational Linguistics,
  Prague, Czech Republic.

\bibitem[{Li and Srikumar(2016)}]{li-srikumar-2016-exploiting}
Li, Tao and Vivek Srikumar. 2016.
\newblock Exploiting sentence similarities for better alignments.
\newblock In \emph{Proceedings of the 2016 Conference on Empirical Methods in
  Natural Language Processing}, pages 2193--2203, Association for Computational
  Linguistics, Austin, Texas.

\bibitem[{Li et~al.(2019)Li, Li, Liu, Meng, and Shi}]{li_word_2019}
Li, Xintong, Guanlin Li, Lemao Liu, Max Meng, and Shuming Shi. 2019.
\newblock On the {Word} {Alignment} from {Neural} {Machine} {Translation}.
\newblock In \emph{Proceedings of the 57th {Annual} {Meeting} of the
  {Association} for {Computational} {Linguistics}}, pages 1293--1303,
  Association for Computational Linguistics, Florence, Italy.

\bibitem[{Liang, Taskar, and Klein(2006)}]{liang-etal-2006-alignment}
Liang, Percy, Ben Taskar, and Dan Klein. 2006.
\newblock Alignment by agreement.
\newblock In \emph{Proceedings of the Human Language Technology Conference of
  the {NAACL}, Main Conference}, pages 104--111, Association for Computational
  Linguistics, New York City, USA.

\bibitem[{Liu et~al.(2016)Liu, Utiyama, Finch, and
  Sumita}]{liu-etal-2016-neural}
Liu, Lemao, Masao Utiyama, Andrew Finch, and Eiichiro Sumita. 2016.
\newblock Neural machine translation with supervised attention.
\newblock In \emph{Proceedings of {COLING} 2016, the 26th International
  Conference on Computational Linguistics: Technical Papers}, pages 3093--3102,
  The COLING 2016 Organizing Committee, Osaka, Japan.

\bibitem[{Liu, Liu, and Lin(2005)}]{liu-etal-2005-log}
Liu, Yang, Qun Liu, and Shouxun Lin. 2005.
\newblock Log-linear models for word alignment.
\newblock In \emph{Proceedings of the 43rd Annual Meeting of the Association
  for Computational Linguistics ({ACL}{'}05)}, pages 459--466, Association for
  Computational Linguistics, Ann Arbor, Michigan.

\bibitem[{Luong, Pham, and Manning(2015)}]{luong-etal-2015-effective}
Luong, Thang, Hieu Pham, and Christopher~D. Manning. 2015.
\newblock Effective approaches to attention-based neural machine translation.
\newblock In \emph{Proceedings of the 2015 Conference on Empirical Methods in
  Natural Language Processing}, pages 1412--1421, Association for Computational
  Linguistics, Lisbon, Portugal.

\bibitem[{Ma and Xia(2014)}]{ma-xia-2014-unsupervised}
Ma, Xuezhe and Fei Xia. 2014.
\newblock Unsupervised dependency parsing with transferring distribution via
  parallel guidance and entropy regularization.
\newblock In \emph{Proceedings of the 52nd Annual Meeting of the Association
  for Computational Linguistics (Volume 1: Long Papers)}, pages 1337--1348,
  Association for Computational Linguistics, Baltimore, Maryland.

\bibitem[{MacCartney, Galley, and
  Manning(2008)}]{maccartneyPhraseBasedAlignmentModel2008}
MacCartney, Bill, Michel Galley, and Christopher~D. Manning. 2008.
\newblock A {{Phrase-Based Alignment Model}} for {{Natural Language
  Inference}}.
\newblock In \emph{Proceedings of the 2008 {{Conference}} on {{Empirical
  Methods}} in {{Natural Language Processing}}}, pages 802--811, {Association
  for Computational Linguistics}, {Honolulu, Hawaii}.

\bibitem[{Mi, Wang, and Ittycheriah(2016)}]{mi-etal-2016-vocabulary}
Mi, Haitao, Zhiguo Wang, and Abe Ittycheriah. 2016.
\newblock Vocabulary manipulation for neural machine translation.
\newblock In \emph{Proceedings of the 54th Annual Meeting of the Association
  for Computational Linguistics (Volume 2: Short Papers)}, pages 124--129,
  Association for Computational Linguistics, Berlin, Germany.

\bibitem[{M{\"u}ller(2017)}]{mullerTreatmentMarkupStatistical2017}
M{\"u}ller, Mathias. 2017.
\newblock Treatment of {{Markup}} in {{Statistical Machine Translation}}.
\newblock In \emph{Proceedings of the {{Third Workshop}} on {{Discourse}} in
  {{Machine Translation}}}, pages 36--46, {Association for Computational
  Linguistics}, {Copenhagen, Denmark}.

\bibitem[{Nagata, Chousa, and Nishino(2020)}]{nagata_supervised_2020}
Nagata, Masaaki, Katsuki Chousa, and Masaaki Nishino. 2020.
\newblock A {Supervised} {Word} {Alignment} {Method} based on
  {Cross}-{Language} {Span} {Prediction} using {Multilingual} {BERT}.
\newblock In \emph{Proceedings of the 2020 {Conference} on {Empirical}
  {Methods} in {Natural} {Language} {Processing} ({EMNLP})}, pages 555--565,
  Association for Computational Linguistics, Online.

\bibitem[{Neubig(2017)}]{Neubig2017NeuralMT}
Neubig, Graham. 2017.
\newblock Neural machine translation and sequence-to-sequence models: A
  tutorial.
\newblock \emph{ArXiv}, abs/1703.01619.

\bibitem[{Nystr{\"o}m et~al.(2006)Nystr{\"o}m, Merkel, Ahrenberg, Zweigenbaum,
  Petersson, and {\AA}hlfeldt}]{nystrom2006creating}
Nystr{\"o}m, Mikael, Magnus Merkel, Lars Ahrenberg, Pierre Zweigenbaum,
  H{\aa}kan Petersson, and Hans {\AA}hlfeldt. 2006.
\newblock Creating a medical english-swedish dictionary using interactive word
  alignment.
\newblock \emph{BMC medical informatics and decision making}, 6(1):1--12.

\bibitem[{Och and Ney(2003)}]{och2003systematic}
Och, Franz~Josef and Hermann Ney. 2003.
\newblock A systematic comparison of various statistical alignment models.
\newblock \emph{Computational linguistics}, 29(1):19--51.

\bibitem[{Papineni et~al.(2002)Papineni, Roukos, Ward, and
  Zhu}]{papineni-etal-2002-bleu}
Papineni, Kishore, Salim Roukos, Todd Ward, and Wei-Jing Zhu. 2002.
\newblock {B}leu: a method for automatic evaluation of machine translation.
\newblock In \emph{Proceedings of the 40th Annual Meeting of the Association
  for Computational Linguistics}, pages 311--318, Association for Computational
  Linguistics, Philadelphia, Pennsylvania, USA.

\bibitem[{Peter, Nix, and Ney(2017)}]{peter2017generating}
Peter, Jan-Thorsten, Arne Nix, and Hermann Ney. 2017.
\newblock Generating alignments using target foresight in attention-based
  neural machine translation.
\newblock \emph{Prague Bull. Math. Linguistics}, 108(1):27--36.

\bibitem[{Radford et~al.(2018)Radford, Narasimhan, Salimans, and
  Sutskever}]{gpt}
Radford, Alec, Karthik Narasimhan, Tim Salimans, and Ilya Sutskever. 2018.
\newblock Improving language understanding by generative pre-training.

\bibitem[{Raganato et~al.(2021)Raganato, V{\'a}zquez, Creutz, and
  Tiedemann}]{raganatoEmpiricalInvestigationWord2021}
Raganato, Alessandro, Ra{\'u}l V{\'a}zquez, Mathias Creutz, and J{\"o}rg
  Tiedemann. 2021.
\newblock An {{Empirical Investigation}} of {{Word Alignment Supervision}} for
  {{Zero-Shot Multilingual Neural Machine Translation}}.
\newblock In \emph{Proceedings of the 2021 {{Conference}} on {{Empirical
  Methods}} in {{Natural Language Processing}}}, pages 8449--8456, {Association
  for Computational Linguistics}, {Online and Punta Cana, Dominican Republic}.

\bibitem[{Rasooli(2019)}]{rasooliCrossLingualTransferNatural2019}
Rasooli, Mohammad~Sadegh. 2019.
\newblock \emph{Cross-{{Lingual Transfer}} of {{Natural Language Processing
  Systems}}}.
\newblock Ph.D. thesis, Columbia University.

\bibitem[{Rasooli and Collins(2015)}]{rasooli-collins-2015-density}
Rasooli, Mohammad~Sadegh and Michael Collins. 2015.
\newblock Density-driven cross-lingual transfer of dependency parsers.
\newblock In \emph{Proceedings of the 2015 Conference on Empirical Methods in
  Natural Language Processing}, pages 328--338, Association for Computational
  Linguistics, Lisbon, Portugal.

\bibitem[{Rasooli and Collins(2017)}]{rasooli-collins-2017-cross}
Rasooli, Mohammad~Sadegh and Michael Collins. 2017.
\newblock Cross-lingual syntactic transfer with limited resources.
\newblock \emph{Transactions of the Association for Computational Linguistics},
  5:279--293.

\bibitem[{Rasooli et~al.(2018)Rasooli, Farra, Radeva, Yu, and
  McKeown}]{rasooli2018cross}
Rasooli, Mohammad~Sadegh, Noura Farra, Axinia Radeva, Tao Yu, and Kathleen
  McKeown. 2018.
\newblock Cross-lingual sentiment transfer with limited resources.
\newblock \emph{Machine Translation}, 32(1):143--165.

\bibitem[{Song et~al.(2020)Song, Wang, Yu, Zhang, Huang, Luo, Duan, and
  Zhang}]{songAlignmentEnhancedTransformerConstraining2020}
Song, Kai, Kun Wang, Heng Yu, Yue Zhang, Zhongqiang Huang, Weihua Luo, Xiangyu
  Duan, and Min Zhang. 2020.
\newblock Alignment-{{Enhanced Transformer}} for {{Constraining NMT}} with
  {{Pre-Specified Translations}}.
\newblock \emph{Proceedings of the AAAI Conference on Artificial Intelligence},
  34(05):8886--8893.

\bibitem[{Stengel-Eskin et~al.(2019)Stengel-Eskin, Su, Post, and
  Van~Durme}]{stengel-eskin-etal-2019-discriminative}
Stengel-Eskin, Elias, Tzu-ray Su, Matt Post, and Benjamin Van~Durme. 2019.
\newblock A discriminative neural model for cross-lingual word alignment.
\newblock In \emph{Proceedings of the 2019 Conference on Empirical Methods in
  Natural Language Processing and the 9th International Joint Conference on
  Natural Language Processing (EMNLP-IJCNLP)}, pages 910--920, Association for
  Computational Linguistics, Hong Kong, China.

\bibitem[{Sutskever, Vinyals, and Le(2014)}]{sutskever2014sequence}
Sutskever, Ilya, Oriol Vinyals, and Quoc~V Le. 2014.
\newblock Sequence to sequence learning with neural networks.
\newblock \emph{Advances in neural information processing systems}, 27.

\bibitem[{Tezcan and Vandeghinste(2011)}]{tezcan-vandeghinste-2011-smt}
Tezcan, Arda and Vincent Vandeghinste. 2011.
\newblock {SMT}-{CAT} integration in a technical domain: Handling {XML} markup
  using pre {\&} post-processing methods.
\newblock In \emph{Proceedings of the 15th Annual conference of the European
  Association for Machine Translation}, European Association for Machine
  Translation, Leuven, Belgium.

\bibitem[{Thadani, Martin, and White(2012)}]{thadani-etal-2012-joint}
Thadani, Kapil, Scott Martin, and Michael White. 2012.
\newblock A joint phrasal and dependency model for paraphrase alignment.
\newblock In \emph{Proceedings of {COLING} 2012: Posters}, pages 1229--1238,
  The COLING 2012 Organizing Committee, Mumbai, India.

\bibitem[{Thadani and McKeown(2011)}]{thadani-mckeown-2011-optimal}
Thadani, Kapil and Kathleen McKeown. 2011.
\newblock Optimal and syntactically-informed decoding for monolingual
  phrase-based alignment.
\newblock In \emph{Proceedings of the 49th Annual Meeting of the Association
  for Computational Linguistics: Human Language Technologies}, pages 254--259,
  Association for Computational Linguistics, Portland, Oregon, USA.

\bibitem[{Vaswani et~al.(2017)Vaswani, Shazeer, Parmar, Uszkoreit, Jones,
  Gomez, Kaiser, and Polosukhin}]{vaswani2017attention}
Vaswani, Ashish, Noam Shazeer, Niki Parmar, Jakob Uszkoreit, Llion Jones,
  Aidan~N. Gomez, Lukasz Kaiser, and Illia Polosukhin. 2017.
\newblock Attention is all you need.
\newblock In \emph{Advances in Neural Information Processing Systems 30: Annual
  Conference on Neural Information Processing Systems 2017, December 4-9, 2017,
  Long Beach, CA, {USA}}, pages 5998--6008.

\bibitem[{Vogel, Ney, and Tillmann(1996)}]{vogel-etal-1996-hmm}
Vogel, Stephan, Hermann Ney, and Christoph Tillmann. 1996.
\newblock {HMM}-based word alignment in statistical translation.
\newblock In \emph{{COLING} 1996 Volume 2: The 16th International Conference on
  Computational Linguistics}.

\bibitem[{Wu et~al.(2005)Wu, Wang, Liu, and Tang}]{wu-etal-2005-improving}
Wu, Hua, Haifeng Wang, Zhanyi Liu, and Kai Tang. 2005.
\newblock Improving translation memory with word alignment information.
\newblock In \emph{Proceedings of Machine Translation Summit X: Posters}, pages
  364--371, Phuket, Thailand.

\bibitem[{Yao(2014)}]{yao2014feature}
Yao, Xuchen. 2014.
\newblock \emph{Feature-driven question answering with natural language
  alignment}.
\newblock Ph.D. thesis, Johns Hopkins University.

\bibitem[{Zenkel, Wuebker, and DeNero(2019)}]{zenkel_adding_2019}
Zenkel, Thomas, Joern Wuebker, and John DeNero. 2019.
\newblock Adding {Interpretable} {Attention} to {Neural} {Translation} {Models}
  {Improves} {Word} {Alignment}.
\newblock \emph{arXiv:1901.11359 [cs]}.
\newblock ArXiv: 1901.11359.

\bibitem[{Zenkel, Wuebker, and DeNero(2020)}]{zenkel_end--end_2020}
Zenkel, Thomas, Joern Wuebker, and John DeNero. 2020.
\newblock End-to-{End} {Neural} {Word} {Alignment} {Outperforms} {GIZA}++.
\newblock In \emph{Proceedings of the 58th {Annual} {Meeting} of the
  {Association} for {Computational} {Linguistics}}, pages 1605--1617,
  Association for Computational Linguistics, Online.

\bibitem[{Zhang et~al.(2017)Zhang, Wang, Liu, and
  Zhou}]{zhangIncorporatingWordReordering2017}
Zhang, Jinchao, Mingxuan Wang, Qun Liu, and Jie Zhou. 2017.
\newblock Incorporating {{Word Reordering Knowledge}} into {{Attention-based
  Neural Machine Translation}}.
\newblock In \emph{Proceedings of the 55th {{Annual Meeting}} of the
  {{Association}} for {{Computational Linguistics}} ({{Volume}} 1: {{Long
  Papers}})}, pages 1524--1534, {Association for Computational Linguistics},
  {Vancouver, Canada}.

\bibitem[{Zhao and Gildea(2010)}]{zhao-gildea-2010-fast}
Zhao, Shaojun and Daniel Gildea. 2010.
\newblock A fast fertility hidden {M}arkov model for word alignment using
  {MCMC}.
\newblock In \emph{Proceedings of the 2010 Conference on Empirical Methods in
  Natural Language Processing}, pages 596--605, Association for Computational
  Linguistics, Cambridge, MA.

\end{thebibliography}

\appendix
\renewcommand{\thesubsection}{\Alph{section}.\arabic{subsection}}
\appendixsection{More on Statistical Machine Translation}
\subsection{Moses: A Statistical Machine Translation System}
\label{appsec:moses}

We now provide further details on the Moses machine translation system\footnote{The full documentation, on which we base this section on, can be found at \url{https://www.statmt.org/moses/}}. Moses learns both \textit{phrase-based} and \textit{linguistic information} about translations. The former involves leveraging co-occurences of phrases, or continuous sequences of words, whereas the latter is to be added externally by knowledgable users. SMT methods which use both of these two are known as \textbf{factored translation}.

The Moses software package consists of a \textbf{training pipeline} and a \textbf{decoder}. The training pipeline is a collection of scripts that takes as input raw data, and outputs a trained MT model. The decoder takes as input a source sentence, and gives as output a ranked list of candidate target translations.

\subsubsection{Moses Training Pipeline}
The training pipeline can further be deconstructed into preprocessing text, obtaining phrase translation tables, and training language models.

\paragraph{Preprocessing text}
In the first step, we convert the parallel corpus into a format suitable for machine use. We read over the bitext to obtain vocabulary files for both source and target languages, which map words to indices. Indices descend from index \texttt{1}, the most common word, to \texttt{T} the $T$-th most common word, where $T$ is the vocabulary size. Words less common than $T$ are assigned the token \texttt{UNK} and the index \texttt{0}, as described in Section~\ref{ssec:tok_vocab}. We then use these vocabulary files to create a new bitext, replacing surface-form tokens with their indices. This bitext is then easily and efficiently machine-readable.

As an additional preprocessing step, we calculate \textit{word classes} for each language---given a word, place it into a set of words with similar usage.

\paragraph{Word alignment and translation tables}
\label{sssec:align_ttables} Word alignment provides the phrase-based translation tables, which are key to the Moses phrase-based translation approach. The aligner used is GIZA++~\cite{och2003systematic}, which we further describe in Section~\ref{ssec:giza}.

First, GIZA++ is trained, in an unsupervised fashion, on a preprocessed parallel corpus. This model then returns word alignments in each direction $F\rightarrow E$ and $E\rightarrow F$. Note that for unsupervised word alignment there are no train/validation/test splits.

Second, we need to combine the $F\rightarrow E$ and $E\rightarrow F$ word alignments. This is done heuristically, the default of which is \texttt{grow-diag-final}. At a high-level, this heuristic begins with the alignment points from the intersection, then adds alignment points from the union if and only if both source and target words for that point were unaligned.

Third, we obtain lexical translation tables by collecting all aligned pairs from the word alignments. The translation tables enumerate words and for each gives its possible translations and associated probability.

Fourth, we extract all phrases into one file, where each line contains the source phrase, translated phrase, and the alignment points.

Fifth, we now build a phrase translation table in the forward direction ($F$ to $E$) using the outputs of the prior two steps. Each line of this table estimates the transition probability $\phi(e|f)$ of phrases. Each line of this table contains a phrase, its candidate translations, and several translation scores. We also build a phrase translation table from $E$ to $F$ similarly, except inverting the phrase translation direction.

\paragraph{Language model}
The language model (LM) is a stage of the training pipeline, which models the likelihood of sentences of a given language, in this case the target language. This allows us to score the different translations of a source sentence.

Moses supports several LMs, using KenLM~\cite{heafield-2011-kenlm}. by default. In brief, these LMs are all n-gram models, which learn from a large monolingual corpus the probability of sequences of words. Typically, 4-grams are used, which means that sentences are decomposed into 4-grams and scored. For more on n-gram language models, see \citet{heafield-2011-kenlm}.

Related to the language model is the \textit{reordering model}. Reordering of phrases is necessary as word order, of course, differs across languages. Details for how this reordering model is learned can be found at \url{http://www2.statmt.org/moses/?n=FactoredTraining.BuildReorderingModel}.

\subsubsection{Moses Decoder}
The Moses decoder is the key step in performing translations. It requires the translation tables and language model from the training pipeline step. We describe its mechanism in brief.  The decoder segments source sentences into source phrases, then translates these phrases into target phrases. The main source of knowledge is the phrase translation tables. There are several possible segmentations, which result in several sets of translated phrases. For each, Moses uses the reordering model to obtain candidate sentence translations. These translations are scored for target language fluency by the language models. As a final heuristic, we impose a word length penalty, to ensure translations are not too much longer or shorter than the source sentences. The intuition is that if too long, a translation likely contains extra information or repetitions, and if too short, a translation likely is missing information.

\subsection{Other Statistical Word Alignment Methods}
\label{appsec:swa}
The fertility-based word aligners of IBM Models 3 to 6 are very slow to train due to their complexity. This is even the case with the computing resources available in the 2020s, given that GIZA++ implementations run only on CPU thread(s). Researchers therefore often turn to these other two word aligners, which are simpler, yet still fairly accurate.

\paragraph{fast\_align} fast\_align~\cite{dyer-etal-2013-simple} is a software package which implements a reparameterized version of IBM Model 2.
Recall Model 1, which has a uniform alignment prior (see Equation~\ref{eq:model1}). Model 2 additionally incorporates an alignment probability $Pr(a_j | j, m, n)$, with the equation
\begin{equation}\label{eq:model}
    Pr(F,a|E) = Pr(m|n) \cdot \prod_{j=1}^{m} Pr(a_j | j, n, m) \cdot Pr(F_j|E_{a_j}) ,
\end{equation}
fast\_align thus trains ten times faster than GIZA++. In practice, fast\_align and GIZA++ have similar performance for language pairs with less reordering (i.e. French and English, both are SOV), and worse performance for those with more reordering (e.g. German is SOV, English is SVO).

\paragraph{Berkeley Aligner} Berkeley Aligner is the software package implementing the HMM-based model of~\citet{liang-etal-2006-alignment}. This work jointly trains two simple HMM models (HMM aligners were described in Section~\ref{ssec:giza}), one in either direction, to agree on their alignments. They report improvements in AER over IBM Model 1 and 2 baselines. In practice, Berkeley Aligner and GIZA++ have similar performance.

\appendixsection{Neural Networks}
\subsection{How Neural Networks Work}
\label{appsec:nn}
A \textbf{neural network} is a machine learning algorithm based on a network of nodes (or neurons) arranged in layers. Each node receives as input \textit{signals} from prior nodes, and compute an output based on a weighted sum of its inputs. The associated \textit{weight} for each node modifies the strength of each signal. A non-linear function is then applied to the output, which then becomes the input signal for the next node(s). 

\begin{figure}[t]
    \centering
    \includegraphics[width=.4\textwidth]{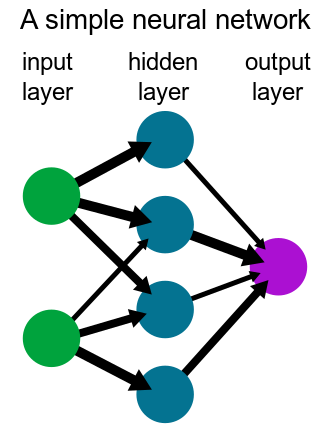}
    \caption{A 3-layer neural network, reproduced from \url{http://en.wikipedia.org/wiki/Neural_network}.}
    \label{fig:nn}
\end{figure}

Figure~\ref{fig:nn} shows a simple neural network, which consists of 3 layers. Each node can be thought of as performing its own linear regression. Another representation of neural networks is as a series of matrix multiplications, with added non-linearity between steps. 

Because of the non-linearity used by neural networks, they are extremely powerful. A neural network with just 3 layers -- input, 1 hidden, output -- can theoretically model any possible function, if the hidden layer is large enough. In practice, many hidden layers are used to learn better representations of problems, especially for the complex domains of images and language.

Furthermore, neural networks are effectively trained, and thus useful, because they model their complicated functions as a chain of simpler functions. All these simpler functions are differentiable. The entire learning algorithm for a neural network is called \textit{automatic differentiation}, which consists of two steps:
\begin{itemize}
    \item \textit{Forward calculation}, passing the input through the neural network, which computes the model output
    \item \textit{Backpropagation}, traversing the neural network backwards, computing the gradient of the loss function (which compares the actual output vs the expected output) with respect to the weights of the network
\end{itemize}

Automatic differentiation, as the name implies, is an efficient way to calculate the gradient. After one step of automatic differentiation, we update the weights of the network with \textit{gradient descent} methods, taking a step towards lower (training) loss, and repeating until convergence. 
The final neural network, and its associated set of weights, serve as an approximate solution the given task.

\subsection{Using Neural Networks}
Due to the representational power of neural networks, they can almost perfectly fit any set of training examples. As we want models to be able to generalize to unseen examples, researchers use several regularization techniques to avoid overfitting to the training data. One such technique is \textit{early stopping}, which involves evaluating the loss of a in-progress model on a held-out validation set, and stopping training when the loss fails to improve after some threshold of steps. Another technique is \textit{dropout}, which randomly zeros out nodes of a neural network during training.

The basic neural network we described above is called a \textit{feedforward neural network} (or multilayer perceptron). Of course, researchers these days use more advanced neural networks. These are especially important to handle temporal sequences of variable length -- such as sequences of words! Two classes of neural networks to handle sequences are \textit{recurrent neural networks} (RNNs) and transformers.
RNNs allows hidden states to reference the outputs of previous hidden states. Transformers use the (self-)\textbf{attention} mechanism, which allow the model to weight previous parts of the input at each step. We further describe attention, and how it correlates to word alignment, in Section~\ref{sec:att}.

\begin{figure}[t]
    \centering
    \includegraphics[width=.9\textwidth]{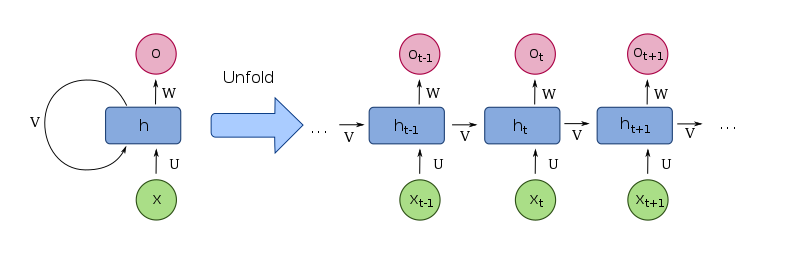}
    \caption{Two views of an RNN. Left: the compressed RNN, Right: the unrolled RNN. Reproduced from \url{http://en.wikipedia.org/wiki/Recurrent_neural_network}.}
    \label{fig:rnn}
\end{figure}

Modern-day neural networks can be very deep (hence the moniker deep learning). This is done by stacking smaller neural networks, and connecting their inputs and outputs. Therefore it is intuitive in diagrams to represent these smaller neural networks as blocks. For example, consider Diagram~\ref{fig:rnn}, a simple recurrent neural network.

\appendixsection{Transformers}
\label{appsec:trans}
Here we summarize the contributions of the transformer~\cite{vaswani2017attention}. First, the use of self-attention as described in Section~\ref{sssec:att_trans}. Second, it uses \textit{multi-head attention}; instead of computing attention once, we compute attention multiple times in parallel. \citet{vaswani2017attention} find that this approach ``allows the model to jointly attend to information from different representation subspaces at different positions.'' Third, they introduce positional encodings, which are summed with word embeddings to preserve positional information in the sentence. Fourth, they stack these self-attention modules, i.e. the encoder consists of stacked self-attention networks, as does the decoder.

Because the transformer removes recurrence, training is highly parallelizable. This allows transformer-based language models to be efficiently trained on on huge amounts of data. To see why this is, let us first consider an RNN. Because the input to each step requires the output of the previous step, we must wait for the previous step to finish. This step requires its previous step, and so on. Each word, therefore, depends on the output of the RNN function, which is highly nonlinear.

In contrast, a transformer uses self-attention. At each decode step, then, we have direct access to the the information to the prior steps---recall the attention mechanism keeps a set of prior hidden states. To generate output at a given time step $t$, we still need to consider the prior $t-1$ steps. But each can be looked up in constant time, meaning we now have only a linear dependence\footnote{See \citet{vaswani2017attention} for further explanation on how self-attention can be parallelized through matrix multiplications.}.

In sum, RNNs require sequential training, given the nonlinear recurrence function. Transformers do not have to be trained sequentially, since attention we can linearly look up the context for any input position. This makes training parallelizable, allowing for efficient training of language models on huge amounts of data.

\appendixsection{Other Neural Word Alignment Methods}
\subsection{Neural Word Alignment, Continued}
\label{appsec:nwa}
In this section, we cover a few neural word aligners that were not included in the main text, as they did not report AER for the German-English word alignment dataset.

\paragraph{Unsupervised}

\citet{cheng2016agreement} propose an approach to jointly train an attentional NMT model to agree on source-to-target and target-to source attention matrices, inspired by~\citet{liang-etal-2006-alignment} (we refer to this as agreement-based training). They report results for a Chinese-English word alignment test set~\cite{liu-etal-2005-log} (we refer to the Chinese to English direction here on as zh-en) of 500 manually aligned sentence pairs, finding the jointly trained model achieves 47.5 AER vs. 52.5 AER for an independently trained model.

\citet{zhangIncorporatingWordReordering2017} explicitly incorporate word reordering knowledge into attentional NMT. Inspired by statistical alignment approaches~\cite{och2003systematic}, they add a word reordering penalty to the calculation of attention. The best model achieves 46.9 AER vs 49.7 on zh-en.

\paragraph{Guided}
\citet{liu-etal-2016-neural} propose to supervise attentional NMT models with alignments from statistical aligners. They convert these (predicted) hard alignments to soft alignments (See Footnote \ref{foot:7}). At training time, they jointly supervise across translation and attention, whereas inference time proceeds as before. Their model achieves 43.4 AER (vs 30.6 GIZA, 50.6 baseline) on zh-en.

\citet{stengel-eskin-etal-2019-discriminative} introduce a discriminative neural alignment model, which measures the dot-product distance between learned source and target vectors. From this distance, the model learns if source-target pairs should be aligned. They further use convolution to consider neighboring decisions. This approach works for any NMT model, but they find best results with a transformer. Trained on a small amount of labeled data, the model achieves 73.4 F1 for a Chinese-English test set\footnote{\url{http://catalog.ldc.upenn.edu/LDC2015T06}} of 636 aligned sentences, vs 62.0 F1 for fast\_align.


\end{document}